\documentclass{article}
\PassOptionsToPackage{numbers, sort&compress}{natbib}
\usepackage[main, final]{neurips_2026}
\usepackage[utf8]{inputenc} 
\usepackage[T1]{fontenc}    
\usepackage{hyperref}       
\hypersetup{
    colorlinks=true,
    citecolor={[RGB]{0,113,188}},
    linkcolor={[RGB]{237,28,36}},
    urlcolor={[RGB]{255,0,128}}
}
\usepackage{url}            
\usepackage{booktabs}       
\usepackage{amsfonts}       
\usepackage{amsmath}        
\usepackage{nicefrac}       
\usepackage{microtype}      
\usepackage{xcolor}         
\usepackage{graphicx}       
\usepackage{enumitem}       
\usepackage{makecell}
\usepackage{booktabs}
\usepackage{amsthm}
\newtheorem{proposition}{Proposition}
\usepackage{amssymb} 
\usepackage{subcaption}
\usepackage{wrapfig}
\usepackage{algorithm,algorithmic}
\usepackage{tabularx}
\usepackage{adjustbox}
\usepackage[table]{xcolor} 
\usepackage{cleveref}
\usepackage{array}      
\usepackage[utf8]{inputenc} 
\usepackage[T1]{fontenc}    
\usepackage{hyperref}       
\usepackage{url}            
\usepackage{booktabs}       
\usepackage{amsfonts}       
\usepackage{nicefrac}       
\usepackage{microtype}      
\usepackage{xcolor}         

\title{What kills $v$-prediction? A Patch-wise PCA Perspective on Pixel-Space Flow Matching}

\author{%
Tianze Luo$^{1,2}$, Haofeng Huang$^{1}$, Yuan Yao$^{1,2}$\thanks{Corresponding author: \texttt{yaoyuanthu@mail.tsinghua.edu.cn}} \\
$^{1}$Tsinghua University, $^{2}$Shanghai Qi Zhi Institute
}

\begin{document}

\maketitle

\begin{abstract}
Latent Diffusion Models dominate generative modeling but face two-stage training complexity and the inherent reconstruction limits of pre-trained autoencoders. Pixel-space Flow Matching offers a simplified end-to-end alternative yet often encounters training difficulties when using the standard $v$-prediction objective. Recent research attributes this difficulty to the off-manifold nature of noised targets, motivating a shift toward $x$-prediction. In this paper, we investigate this explanation through patch-wise Principal Component Analysis (PCA) and identify a bandwidth bottleneck. The velocity in minor components is largely determined by the raw noisy input, requiring deep backbones to preserve information that competes with principal-component modeling. Through controlled experiments and representation analysis, we further examine why $x$-prediction remains effective: its parameterization provides a raw-input bypass, while its learned input representation can spontaneously concentrate on principal directions. Under low-rank observations, our analysis of the induced flow characterizes $x$-prediction as an implicit conditional-mean decoder, reconstructing discarded components from the retained information. These observations motivate the Raw-injected Transformer (RiT), which combines a lightweight raw-conditioned decoder with an output-side shortcut to separate semantic modeling from raw-state propagation. Our method restores effective $v$-prediction and achieves competitive performance on ImageNet $256\times256$, providing a perspective on the interaction between prediction objectives and architectural bandwidth. Our code is available at {\small \href{https://github.com/THUMAI-Lab/RiT}{\texttt{https://github.com/THUMAI-Lab/RiT}}}.

\end{abstract}

\vspace{-2mm}
\begin{figure}[!ht]
    \centering

    \begin{minipage}[b]{0.67\linewidth}
        \centering
        \includegraphics[width=\linewidth]{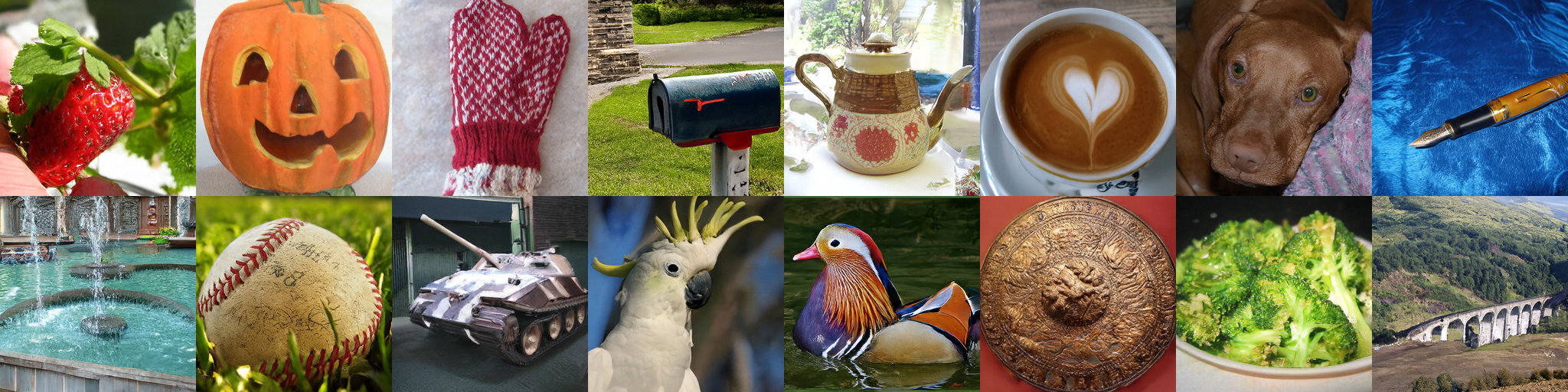}
        \par\vspace{1mm}\nointerlineskip
        {\small\strut (a) Curated samples from RiT-H/16 on ImageNet $256\times256$.}
    \end{minipage}
    \hfill
    \begin{minipage}[b]{0.30\linewidth}
        \centering
        \normalsize
        \setlength{\tabcolsep}{4pt}
        \renewcommand{\arraystretch}{1.26}
        \begin{tabular}{lcc}
            \toprule
            Size & JiT & RiT \\
            \midrule
            B/16 & 3.66 & \textbf{3.01} \\
            L/16 & 2.36 & \textbf{2.06} \\
            H/16 & 1.86 & \textbf{1.79} \\
            \bottomrule
        \end{tabular}
        \par\vspace{1mm}\nointerlineskip
        {\small\strut (b) FID-50K comparison ($\downarrow$)}
    \end{minipage}
    \vspace{2mm}
    \caption{\textbf{Qualitative results and quantitative comparison.} The visual examples illustrate the generation quality of our model, while the table compares FID-50K across different model scales on ImageNet $256\times256$. RiT-B and RiT-L are trained for 600 epochs, whereas RiT-H is trained for 480 epochs. All JiT~\citep{li2025back} baselines are trained for 600 epochs. Our models consistently achieve lower FID-50K scores, even with fewer training epochs at the H scale.}
    \label{fig:teaser_results}
\end{figure}
\vspace{-3mm}

\section{Introduction}
\vspace{-2mm}
Diffusion and Flow Matching Models \citep{ho2020denoising,song2021score,song2019generative,lipman2022flow,liu2023flow,albergo2023building} have emerged as a robust and powerful paradigm for visual generative modeling, demonstrating competitive generation quality and distribution coverage compared to Normalizing Flows and Generative Adversarial Networks (GANs) \citep{salimans2016improved,goodfellow2020generative,brock2019large,kang2023scaling,dhariwal2021diffusion,rombach2022high,qin2025lumina,cai2025z,zhai2025normalizing,ho2019flow++,rezende2015variational,geng2025meanflow,luo2026soflow,frans2024one}. While initially applied to image synthesis, this framework has been successfully extended to diverse high-dimensional domains, including video generation \citep{kong2024hunyuanvideo,wan2025wan,yang2025cogvideox} and audio generation \citep{du2024cosyvoice,cheng2025mmaudio,luo2025wavefm,lee2025periodwave,chen2025f5}. Although diffusion models originally operated in raw pixel space, their performance and training efficiency often degrade as image resolution increases. These difficulties arise in part from directly learning high-frequency spatial dependencies and the quadratic complexity of dense attention mechanisms commonly used in modern diffusion backbones.

To mitigate these computational constraints, Latent Diffusion Models (LDMs) \citep{rombach2022high} have become the dominant approach in recent years. By leveraging pre-trained Variational Autoencoders (VAEs) \citep{kingma2013auto} to project inputs into a compressed, low-dimensional representation, LDMs effectively decouple generative modeling from high-resolution synthesis \citep{peebles2023scalable,ma2024sit}. However, the reliance on complex two-stage pipelines and the reconstruction ceiling of fixed decoders have recently spurred widespread research interest in end-to-end pixel-space diffusion models \citep{chen2025dip,li2025back,yu2025pixeldit}. To maintain computational efficiency, these models typically adopt large patch sizes (e.g., $16\times16$ or $32\times32$), creating high-dimensional patch representations that pose new challenges for standard diffusion objectives.

Crucially, the recent JiT study \citep{li2025back} identifies a severe training failure in pixel space: while velocity prediction ($v$-prediction) is Flow Matching's native regression objective, applying it directly to high-dimensional pixel patches can fail. JiT therefore adopts clean-data prediction ($x$-prediction), which substantially improves generation quality in this setting. However, the cause of this failure remains incompletely understood. This difference is often attributed to the low-dimensional manifold nature of natural images. Yet the MSE-optimal $x$-predictor estimates a conditional expectation, which need not lie on the data manifold. The manifold interpretation alone therefore leaves open how prediction objectives interact with the information constraints of practical architectures.

\begin{figure}[t]
    \centering
    \includegraphics[width=0.92\linewidth]{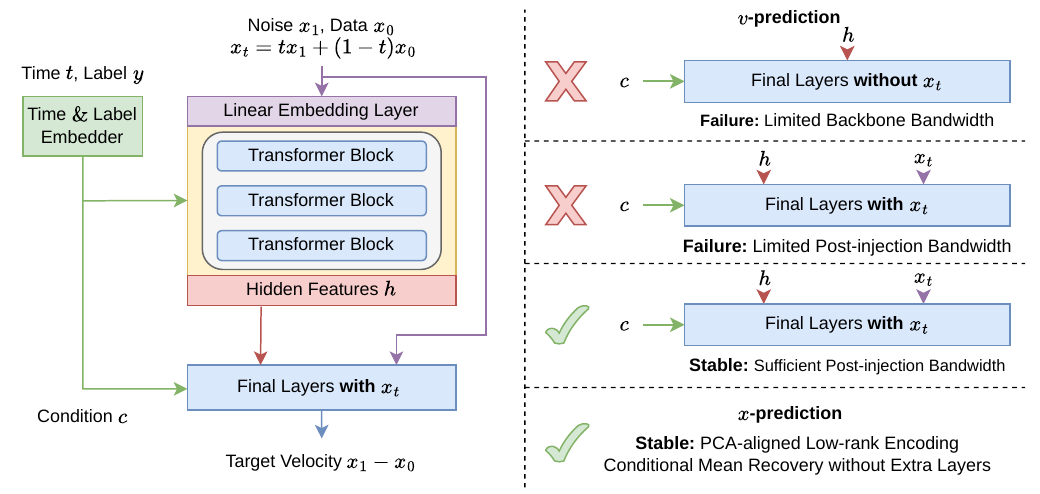}
    \vspace{-1mm}
    \caption{\textbf{Overview of RiT and the bandwidth bottleneck.} Standard $v$-prediction must retain noisy input while learning semantic features, creating competing demands on backbone bandwidth. Enlarging the backbone can alleviate this conflict without extra raw injection, but incurs substantial computational overhead. RiT instead uses a lightweight raw-conditioned nonlinear decoder with sufficient post-injection capacity and an output-side shortcut. Details of the final layers are given in Section~\ref{sec:rit}. In contrast, $x$-prediction provides an inherent raw-input bypass and can learn compact representations concentrated on principal directions. Under low-rank conditions, its induced flow reconstructs discarded components as conditional means without additional decoder layers.}
    \label{fig:rit_overview}
    \vspace{-10mm}
\end{figure}

We investigate this question by analyzing image patches through Principal Component Analysis (PCA). We observe that $v$-prediction converges in low-dimensional principal subspaces but deteriorates as minor components are introduced. Linear reconstruction probing reveals that $v$-prediction retains substantially more raw noisy-input information through the backbone than $x$-prediction. Our affine analysis links this behavior to the raw-state transport required by noise-dominated minor components, which competes with principal-component modeling. The $x$-prediction parameterization provides a raw-input bypass that alleviates this demand. Under low-rank observations, its induced flow completes discarded coordinates through conditional means without propagating their noisy states through the backbone. We further observe that even without an imposed low-rank bottleneck, its input embedding can spontaneously reduce its effective rank and concentrate on principal directions.

To resolve this feature bandwidth conflict, we propose the Raw-injected Transformer (\textbf{RiT}) for pixel-space Flow Matching. RiT injects the raw noisy input $x_t$ into a lightweight subpatch decoder and combines a semantic branch with an explicit raw-input shortcut through a shared output projection. The decoder provides sufficient post-injection bandwidth to model conditional variation beyond mean reconstruction, while the shortcut supplies learnable raw-state transport. Together, these components allow content-dependent processing of the full noisy input without requiring the deep backbone to preserve every coordinate. The final architecture requires neither a precomputed PCA basis nor auxiliary perceptual, adversarial, or frequency-domain losses.

We conduct extensive experiments on the ImageNet $256\times256$ \citep{deng2009imagenet} dataset at B, L, and H scales. Controlled interventions on backbone capacity, patch dimensionality, raw-input availability, and decoder bandwidth support the bandwidth interpretation. RiT restores effective $v$-prediction and achieves competitive FID scores compared to $x$-prediction baselines across training horizons. These results indicate that the observed failure is not an inherent limitation of velocity prediction, but depends on whether the architecture supplies the information paths required by the objective.

\section{Related Work}

\paragraph{Latent Diffusion Models}
Latent diffusion has become a standard paradigm for efficient and scalable image and video generation~\citep{yu2025representation,esser2024scaling,ma2024sit,qin2025lumina,cai2025z,rombach2022high}. Early systems use VAEs~\citep{kingma2013auto} for compact representations, while the transition from U-Net~\citep{ronneberger2015u} to DiT~\citep{peebles2023scalable} improves scalability and generation quality. Subsequent autoencoder designs explore higher compression ratios, channel-wise latent structures, and semantic alignment for downstream diffusion~\citep{chen2024deep,chen2025dc,yao2025reconstruction,wan2025wan}. RAE~\citep{zheng2025diffusion} pairs pretrained representation encoders such as DINOv2~\citep{oquab2023dinov2} with ViT decoders~\citep{dosovitskiy2021image}, forming Transformer-based autoencoders with high-dimensional semantic latents and improving diffusion convergence and generation quality. However, autoencoder training often involves perceptual and adversarial losses, adding optimization overhead and complexity to the two-stage generation pipeline.

\paragraph{Pixel Diffusion Models}
Diffusion models were originally developed in pixel space~\citep{ho2020denoising,song2021score,song2019generative}. Early high-resolution approaches include diffusion upsampling in ADM~\citep{dhariwal2021diffusion} and progressive super-resolution cascades in CDM~\citep{ho2022cascaded}. SiD~\citep{hoogeboom2023simple} and SiD2~\citep{hoogeboom2025simpler} improve pixel-space generation through U-ViT architectures, adapted noise schedules, and loss weighting. PixelFlow~\citep{chen2025pixelflow} trains a unified DiT for multi-resolution Flow Matching, whereas RelayDiffusion~\citep{teng2024relay} uses separate models for different scales. Inspired by implicit neural fields for scene reconstruction~\citep{mildenhall2021nerf,sitzmann2020implicit}, PixNerd~\citep{wang2025pixnerd} introduces neural-field decoding after the DiT backbone, achieving competitive results at patch size 16.

More recently, JiT~\citep{li2025back} finds clean-image prediction more effective than direct velocity prediction for large-patch pixel-space Transformers and attributes this advantage to the low-dimensional data manifold. Other methods augment DiT for high-frequency detail modeling: PixelDiT~\citep{yu2025pixeldit} adds pixel-level Transformer blocks with linear token compression and expansion; DeCo~\citep{ma2025deco} couples a lightweight pixel decoder with a frequency-aware Flow Matching loss; and DiP~\citep{chen2025dip} introduces a lightweight convolutional U-Net Patch Detailer Head. DDT~\citep{wang2026ddt} also separates semantic encoding from velocity decoding by supplying the noisy input directly to a dedicated decoder, but operates in compressed VAE latent space rather than on high-dimensional pixel patches.

\section{Preliminary: Flow Matching}

We review the principles of Flow Matching, a generative framework that transports a tractable prior to a target data distribution.  Let $p(x_0)$ denote the data distribution and $p(x_1)$ represent the prior, typically defined as a standard Gaussian $\mathcal{N}(0,I)$. $x_1$ is sampled independently of $x_0$. The transition between a data sample $x_0$ and a noise sample $x_1$ is defined via linear interpolation, creating a time-dependent state $x_t = (1 - t)x_0 + tx_1 \in \mathbb{R}^n$ for $t \in [0, 1]$. This probability path is governed by a marginal velocity field, given by the conditional expectation of the vector from $x_0$ to $x_1$:
\begin{equation}
    \label{fm}
    v(x_t,t)=\mathbb{E}[x_1-x_0\mid x_t,t].
\end{equation}
To generate data samples, we first sample $x_1\sim p(x_1)$ and then solve the corresponding ordinary differential equation (ODE) backward from $t=1$ to $t=0$:
\begin{equation}
    \frac{\mathrm{d}X(t)}{\mathrm{d}t}=v(X(t),t),\qquad X(1)=x_1.
\end{equation}
In practice, the velocity field is parameterized by a neural network $v_\theta$, and numerical methods such as Euler or Heun are used to approximate the trajectory. Direct regression against the marginal field is generally intractable because its conditional expectation is unavailable. The Conditional Flow Matching formulation~\citep{lipman2022flow,liu2023flow} provides a tractable objective with the same optimization gradients by using individual data-noise pairs:
\begin{equation}
    \mathcal{L}_{\mathrm{CFM}}(\theta)
    =\mathbb{E}_{t,x_0,x_1}\left\|v_\theta(x_t,t)-(x_1-x_0)\right\|_2^2.
    \label{eq:cfm}
\end{equation}
Its optimal predictor recovers the marginal velocity field. Recent works~\citep{li2025back,ma2026pixelgen} demonstrate that direct data prediction ($x$-prediction) is beneficial for pixel-space generation. Under this formulation, the network $\hat x_\theta$ regresses $x_0$, and the objective is re-weighted as
\begin{equation}
    \mathcal{L}_{\mathrm{Data}}(\theta)
    =\mathbb{E}_{t,x_0,x_1}\left[1/\bar{t}^{\,2}\, \|\hat x_\theta(x_t,t)-x_0\|_2^2\ \right],
    \qquad \bar t=\max(t,0.05).
    \label{eq:xloss}
\end{equation}
The time-dependent denominator is thresholded to prevent numerical instability near $t=0$. Without clipping, the velocity is recovered through $v_\theta(x_t,t)=(x_t-\hat x_\theta(x_t,t))/t$ for $t>0$. Our implementation uses $\bar t$ for the denominator in this expression as well. We distinguish the unclipped continuous-time relation from its practical implementation in the analysis below. Class conditioning is omitted here for clarity and denoted by $\ell$ where needed.

\section{Methodology}
\subsection{Overview}
We present our methodology for reviving $v$-prediction in pixel-space Flow Matching. Despite its simple, time-independent regression target $x_1-x_0$, direct $v$-prediction can fail when high-dimensional pixel patches must be processed within a limited hidden representation. We investigate this difficulty through patch-wise PCA, which shows that $v$-prediction trains successfully in lower-dimensional principal subspaces but deteriorates as minor components are introduced. Linear reconstruction probing then reveals a stronger tendency to retain noisy-input information throughout the backbone.

We explain this behavior through an affine analysis of component-wise velocity transport and examine two complementary advantages of $x$-prediction: its raw-input bypass and its tendency to learn representations concentrated on principal directions. Under low-rank observations, we characterize how the induced flow reconstructs discarded components as conditional means. Building on these insights, RiT combines a lightweight raw-conditioned nonlinear decoder with an output-side shortcut. Controlled experiments on backbone width, patch dimensionality, decoder capacity, and raw-input availability test the bandwidth interpretation and clarify the roles of these components.

\subsection{Patch-wise Principal Components Analysis}
\label{sec:pca}
To understand the failure of $v$-prediction in pixel space, we first investigate the distribution of image patches. Natural images are highly redundant, and much of their centered variation is concentrated in a low-dimensional subspace. Let $\mu=\mathbb{E}[x_0]\in\mathbb{R}^p$ and $\Sigma=\operatorname{Cov}(x_0)\in\mathbb{R}^{p\times p}$ denote the patch mean and covariance, where $p$ is the patch dimension. For a patch size of 16, $p=3\times16\times16=768$. By performing eigendecomposition $\Sigma=V\Lambda V^\top$, we obtain the orthonormal eigenbasis $V=[v_1,\ldots,v_p]$ and $\Lambda=\operatorname{diag}(\lambda_1,\ldots,\lambda_p)$, with $\lambda_1\ge\cdots\ge\lambda_p\ge0$. Leading eigenvectors capture the dominant centered variation, while the remaining lower-variance directions are referred to as minor components. We define the affine projection onto the principal $k$-dimensional subspace as
\begin{equation}
    P_k=V_kV_k^\top,\qquad
    \mathcal{P}_k^\mu(x)=\mu+P_k(x-\mu),
    \label{eq:affine-pca}
\end{equation}
where $V_k$ consists of the first $k$ principal eigenvectors. We apply this map to both data and noise, so that both endpoints lie in $\mu+\operatorname{range}(P_k)$. This changes the stochastic dimensionality while preserving the original patch shape, and the map becomes the identity when $k=p$. Our implementation follows the JiT-B backbone~\citep{li2025back}, with hidden size 768 and approximately 132M parameters in these experiments. The intermediate width of its two-layer linear input embedding is also 768, so no architectural low-rank bottleneck is imposed here.

All experiments are conducted on ImageNet~\citep{deng2009imagenet} at $256\times256$ resolution with patch size 16. We employ AdamW with learning rate $2\times10^{-4}$, zero weight decay, and betas $(0.9,0.95)$. Training lasts 250,000 updates with global batch size 1024, including a linear warmup over the initial 6250 updates and a constant learning rate thereafter. Evaluation uses an Exponential Moving Average (EMA) decay of 0.9996 and Fr\'echet Inception Distance (FID-50K)~\citep{heusel2017gans}. We use 50-step Heun sampling with a final Euler step, corresponding to 99 velocity evaluations. Classifier-Free Guidance (CFG)~\citep{ho2022classifier} is disabled to examine generation quality without guidance.

\begin{table}[ht]
  \centering
  \setlength{\tabcolsep}{3.4pt}
    \begin{tabular}{l ccccc ccccc}
    \toprule
    & \multicolumn{5}{c}{$v$-prediction } & \multicolumn{5}{c}{$x$-prediction} \\
    \cmidrule(lr){2-6} \cmidrule(lr){7-11}
    \makecell{PCA Dim $k$} & 128 & 192 & 256 & 384 & 768 (Full) & 128 & 192 & 256 & 384 & 768 (Full) \\
    \midrule
    FID-50K ($\downarrow$) & \textbf{33.99} & 36.49 & 38.75 & 47.41 & 149.07 (Fail) & 35.01 & 37.56 & 39.58 & 40.83 & 40.39 \\
    \bottomrule
\end{tabular}
    \vspace{2mm}
    \caption{\label{tab:pca} \textbf{Effect of PCA projection.}
    While $v$-prediction fails in the full 768-dimensional pixel space, restricting both data and noise to lower-dimensional principal subspaces not only restores effective training but even allows $v$-prediction to outperform $x$-prediction at $k=128,192,256$. Its performance then deteriorates as additional minor directions are introduced.}
\end{table}
\vspace{-4mm}

As reported in Table~\ref{tab:pca}, the two objectives behave differently as the subspace dimension increases. In the 128-dimensional principal subspace, $v$-prediction converges successfully and achieves lower FID than $x$-prediction. At $k=128,192,256$, training in the principal subspace also improves FID relative to full-space training for both objectives at this training budget. However, $v$-prediction is more sensitive to minor components: as the dimension expands from 128 to 384, its FID rises from 33.99 to 47.41, and full-space training deteriorates to 149.07. In contrast, $x$-prediction remains effective across all tested dimensions, reaching an FID of 40.39 at full rank. These observations suggest that $x$-prediction has an advantage in handling noise-dominated minor directions, and that the failure of direct $v$-prediction depends on the information demands placed on the architecture rather than solely on the velocity target itself.

\subsection{The Bandwidth Bottleneck and Noisy Input Memorization}
\label{sec:probe}
\begin{figure}[ht]
    \centering
    \begin{minipage}[c]{0.54\linewidth}
        \centering
        \includegraphics[width=\linewidth]{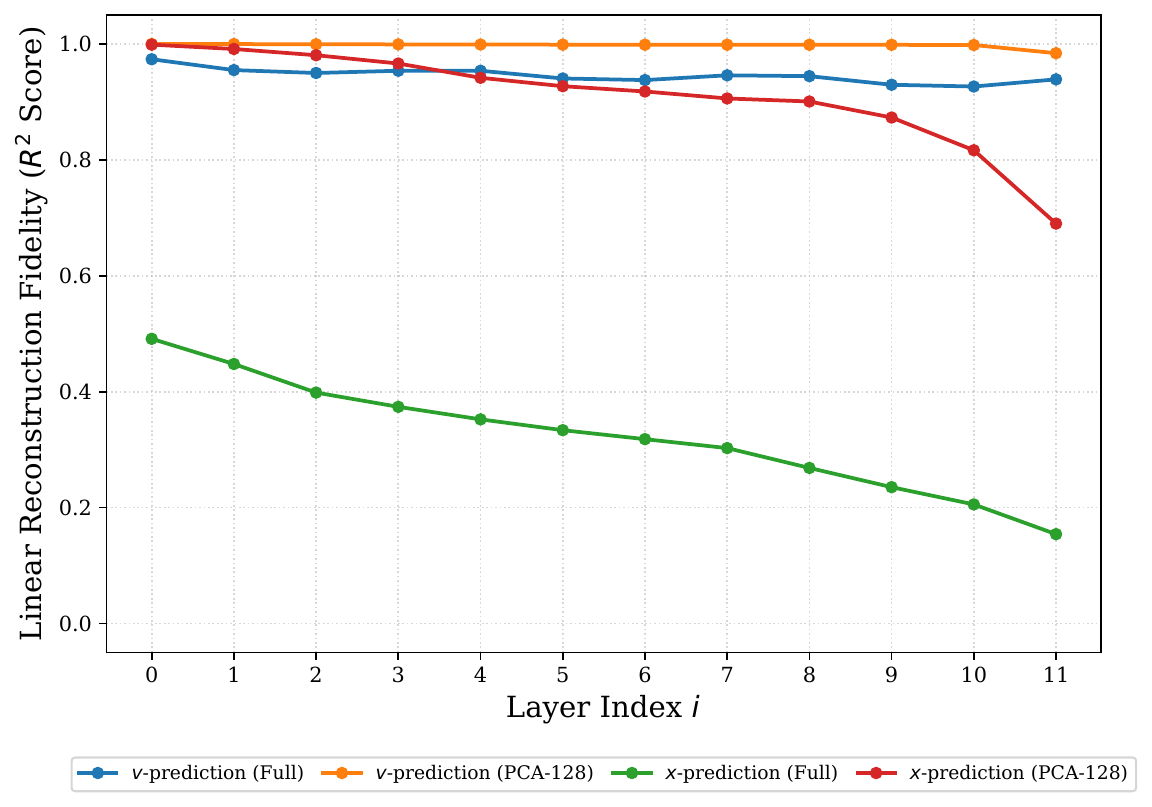}
    \end{minipage}
    \hfill
    \begin{minipage}[c]{0.44\linewidth}
    \vspace{3mm}
    \caption{\textbf{Linear reconstruction fidelity ($R^2$ score) of $x_t$ from $h_l$.} We observe that $v$-prediction tends to retain substantially more noisy-input information in the residual path than $x$-prediction, consistent with the raw-state contribution required for minor-component velocity estimation. When the patch and hidden dimensions are comparable, preserving this 768-dimensional state can compete with the capacity needed for principal-component modeling. In contrast, the PCA-128 input has fewer stochastic dimensions, allowing high input retention alongside effective generative learning.}
        \label{fig:r2_score}
    \end{minipage}
    \vspace{-4mm}
\end{figure}

To uncover why processing minor components leads to difficulty, we analyze information flow within the network. Specifically, we investigate whether hidden states in deep layers retain the noisy input $x_t$. We perform linear probing to reconstruct $x_t$ from the hidden states $h_l$ at each layer $l$, measuring fidelity through the coefficient of determination ($R^2$). After training in the previous section, we fit one affine map per layer on a fixed set of analysis batches:
\begin{equation}
    R^2(h_l,x_t)=1-
    \frac{\min_{W,b}\sum_n\|x_{t,n}-(Wh_{l,n}+b)\|_2^2}
         {\sum_n\|x_{t,n}-\bar x_t\|_2^2},
    \label{eq:r2}
\end{equation}
where $n$ indexes image patches across the batches, and $\bar x_t$ is the empirical mean of the noisy input patches. All models are evaluated on the same images and noise draws, with the corresponding PCA projection applied to subspace models. Figure~\ref{fig:r2_score} reveals that $v$-prediction retains substantially more input information than $x$-prediction. This behavior is consistent with its target $x_1-x_0$, which requires the current noisy state to remain available for minor-component velocity estimation.

\vspace{-4mm}

In full pixel space, $v$-prediction maintains high reconstruction fidelity, with $R^2\approx0.95$ across much of the backbone. When the input dimensionality is comparable to the hidden width, preserving this high-dimensional state can compete with semantic modeling. In contrast, the linear accessibility of $x_t$ decreases with depth under $x$-prediction. Models trained in the PCA-128 subspace also show high input retention, yet do not fail: their 768-dimensional hidden states can accommodate a substantially smaller stochastic input while retaining capacity for nonlinear modeling. These probes indicate that substantially more noisy-input information remains linearly accessible in the hidden states under $v$-prediction, consistent with the increased raw-state routing demands associated with minor components. This connects the degradation in Section~\ref{sec:pca} to a bandwidth conflict between preserving high-dimensional noisy state and modeling principal components, which is further supported by the width, patch-size, and raw-access interventions in Section~\ref{sec:rit}.

\subsection{The Implicit Advantage of \texorpdfstring{$x$}{x}-Prediction on Minor Components}
\label{sec:xadvantage}
Sections~\ref{sec:pca} and~\ref{sec:probe} show that $x$-prediction remains effective while retaining less noisy-input information. We now examine how its parameterization reduces the burden of raw-state propagation and permits reconstruction from low-rank observations. Throughout, $\widetilde x_t=x_t-(1-t)\mu$ and $\mu_i=v_i^\top\mu$ account for the patch mean.

\begin{proposition}[Affine raw-state transport]
\label{prop:affine}
For fixed $t>0$, the minimum-MSE affine estimator of $v_i^\top(x_1-x_0)$ from $v_i^\top x_t$ is $\widehat V_i^{\mathrm{aff}}=-\mu_i+C_i(t)v_i^\top\widetilde x_t$, with
\begin{equation}
    C_i(t)=\frac{t-(1-t)\lambda_i}{t^2+(1-t)^2\lambda_i},\qquad
    \operatorname{std}(r_i(t))=\frac{\sqrt{\lambda_i}}{\sqrt{t^2+(1-t)^2\lambda_i}},
    \label{eq:linear-coef}\vspace{-1mm}
\end{equation}
where $r_i(t)=\widehat V_i^{\mathrm{aff}}-v_i^\top(x_1-x_0)$ measures error against the paired training target.\vspace{-1mm}
\end{proposition}

\begin{figure}[ht]
    \centering
    \begin{minipage}[c]{0.53\linewidth}
        \centering
        \includegraphics[width=\linewidth]{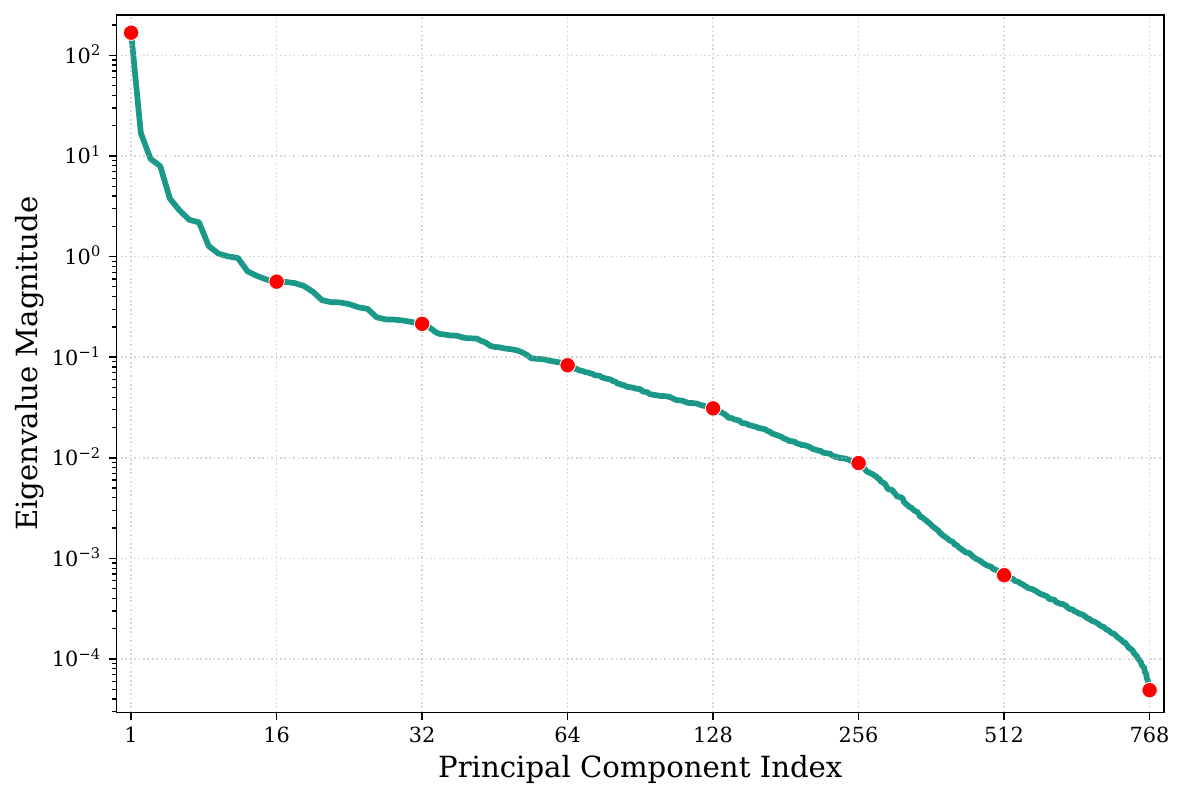}
    \end{minipage}
    \hfill
    \begin{minipage}[c]{0.44\linewidth}
        \vspace{2mm}
        \caption{\textbf{Eigenvalue distribution of image patches.} Eigenvalues of the centered $768\times768$ patch covariance on the ImageNet $256\times256$ training set, shown in descending order. Their magnitudes span more than six orders, reflecting a strongly anisotropic distribution. Minor directions contain little data variance but receive unit-variance Gaussian noise in full-space training. Predicting their velocity therefore requires access to the noisy state even when the remaining regression error is small, motivating a raw-input shortcut that reduces this transmission burden.}
        \label{fig:pca_eigenvalues}
    \end{minipage}
    \vspace{-4mm}
\end{figure}

Figure~\ref{fig:pca_eigenvalues} shows that many minor directions have very small eigenvalues. At fixed $t>0$, $C_i(t)\to1/t$ and the residual variance vanishes as $\lambda_i\to0$, so the affine estimate approaches $(v_i^\top x_t-\mu_i)/t$. In these low-variance directions, the MSE-optimal clean-data prediction is close to the mean $\mu_i$. Thus, the $x$-prediction parameterization $\hat v_\theta=(x_t-\hat x_\theta)/t$ provides approximately the same affine transport away from the low-noise endpoint. This reduces the backbone's transmission burden and motivates a learnable shortcut for $v$-prediction in Section~\ref{sec:rit}. Appendix~\ref{appa:proof} provides the derivation. Importantly, the full $x$-prediction identity remains exact beyond this small-variance approximation.

\vspace{-3mm}

This affine advantage does not fully explain why $x$-prediction remains effective with low-rank linear input embeddings, such as the 128-dimensional bottlenecks used for 768-dimensional patches in JiT-B/16 and JiT-L/16~\citep{li2025back}. Unlike the PCA intervention in Section~\ref{sec:pca}, the full data and noise distributions are retained, but at least 640 directions per patch are unobserved by the backbone. The following result explains how $x$-prediction completes these discarded clean-data coordinates.

\begin{proposition}[Conditional-mean completion]
\label{prop:conditional-mean}
Let $P_1$ project orthogonally onto the row space of the linear input map $W$, and let $P_2=I-P_1$ project onto its nullspace. Suppose the predictor observes only $Y_t=P_1\widetilde x_t$, time, and class $\ell$. Assume the retained flow is learned correctly and has a unique trajectory with an endpoint, and the minor conditional mean extends continuously along it to $t=0$. The MSE-optimal $x$-predictor with the unclipped velocity parameterization then satisfies
\begin{equation}
    P_2\widehat X(0)=\mathbb E[P_2x_0\mid P_1x_0=P_1\widehat X(0),\ell].
    \label{eq:minor-endpoint}
\end{equation}
\end{proposition}   

\begin{figure}[h]
    \centering
    \includegraphics[width=0.49\linewidth]{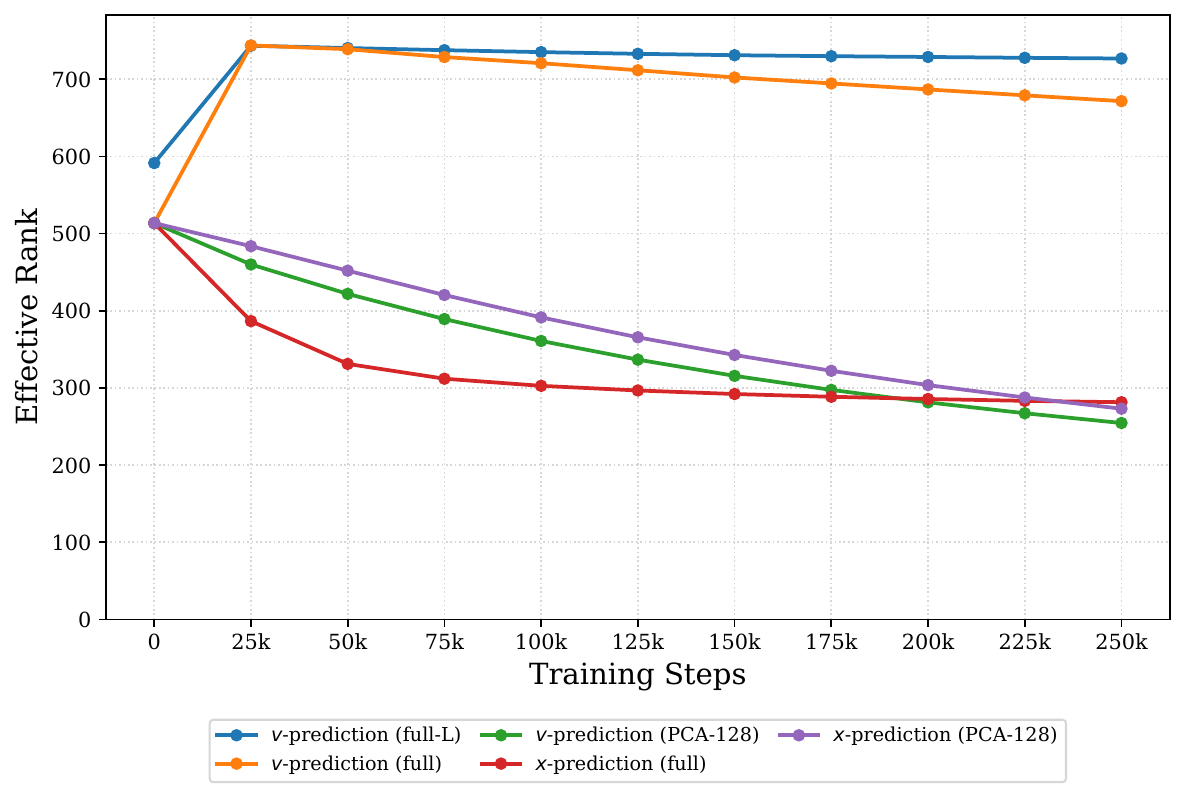}\hfill
    \includegraphics[width=0.49\linewidth]{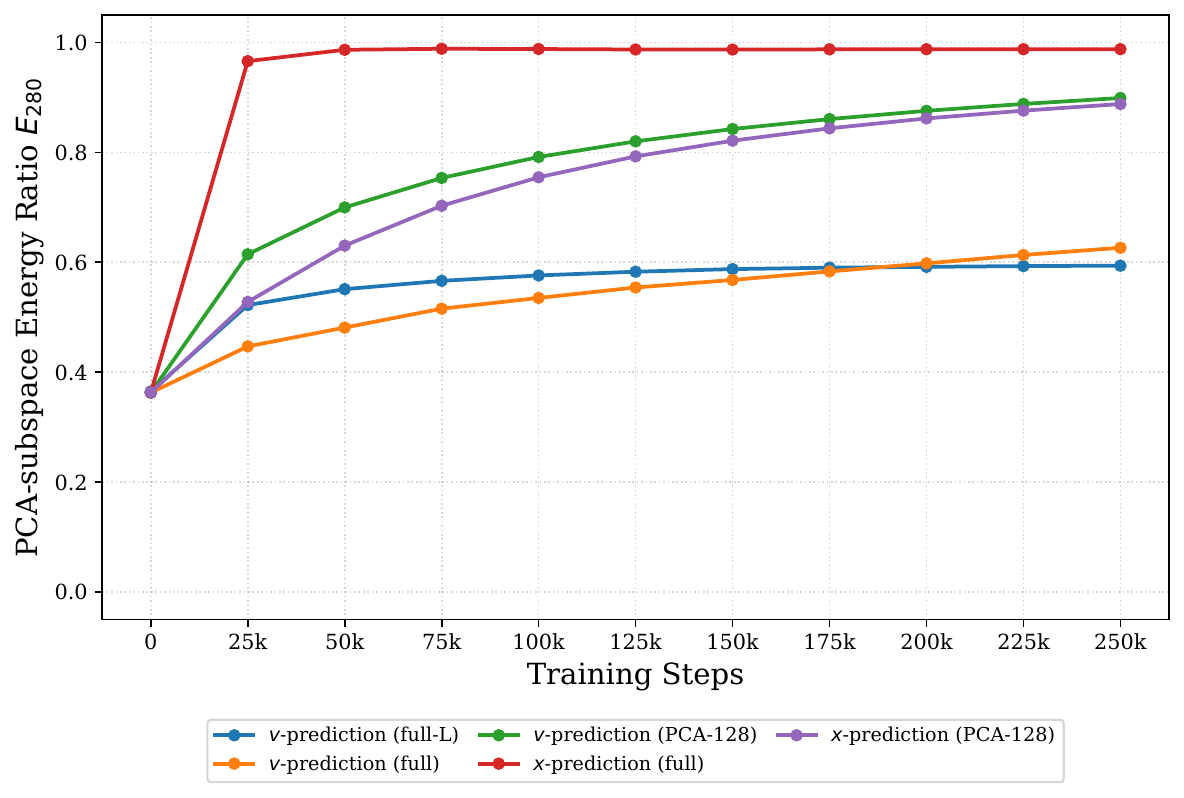}
    \caption{\textbf{Evolution of the learned input embedding.} Left: effective rank of $W=W_2W_1$, computed from the entropy of its normalized singular values to characterize the dimensionality of the learned representation. Right: energy in the first 280 PCA directions. Full-space $x$-prediction concentrates on principal directions, whereas $v$-prediction retains a broader spectrum. PCA-128 denotes projected data and noise, not an embedding bottleneck.}
    \label{fig:embedding}
    \vspace{-2mm}
\end{figure}

The integrating-factor proof in Appendix~\ref{appb:lowrank} shows that $x$-prediction removes the initial noise in unobserved directions and completes their clean-data values through conditional means, which may depend nonlinearly on retained information. Direct $v$-prediction with the same observation instead retains unit-variance minor noise. The appendix also analyzes the effect of time clipping along a fixed retained trajectory. Given its ability to complete discarded coordinates, we next examine whether $x$-prediction also learns more compact input representations without an explicit input bottleneck. Figure~\ref{fig:embedding} shows that its full-width input embedding spontaneously develops a lower effective rank and aligns with principal directions. For the composed map $W=W_2W_1$, we quantify this behavior through its singular-value entropy effective rank~\citep{roy2007effective} and its operator energy in principal directions:
\begin{equation}
    \operatorname{erank}(W)=\exp\!\left(-\sum_i\pi_i\log\pi_i\right),\quad
    \pi_i=\frac{\sigma_i(W)}{\sum_j\sigma_j(W)},\quad
    E_K(W)=\frac{\|WV_K\|_F^2}{\|W\|_F^2}.
    \label{eq:embedding-metrics}
\end{equation}
We fix $K=280$ to approximately match the effective rank of full-space $x$-prediction after 200 epochs and assess PCA alignment using unit-norm PCA directions without eigenvalue weighting. Figure~\ref{fig:embedding} shows that even with a full-width input embedding, $x$-prediction spontaneously reduces its effective rank and concentrates on principal directions, whereas $v$-prediction retains a broader spectrum. Reduced effective rank does not itself imply exact information loss, but this observation supports the relevance of the low-rank analysis.

Conditional-mean completion captures nonlinear dependencies on retained information, but does not generally recover the joint conditional distribution of discarded components. This motivates a lightweight nonlinear decoder with direct raw-input access and sufficient bandwidth to model the remaining conditional variation. Combined with the learnable shortcut, this design improves FID over the $x$-prediction baseline in our experiments, as detailed in Section~\ref{sec:rit}.

\subsection{Raw-Conditioned Decoding Revives \texorpdfstring{$v$}{v}-Prediction}
\label{sec:rit}
Following this analysis, RiT combines a lightweight raw-conditioned decoder for modeling conditional variation with a learnable output shortcut for raw-state transport under direct $v$-prediction. Let $s_b$ and $s_d$ denote the backbone and decoder patch sizes, respectively, with $s_b/s_d$ an integer.

\paragraph{Decoder bandwidth and raw-input access.}
Each backbone token is linearly expanded and rearranged into $(s_b/s_d)^2$ decoder tokens. Let $h^{\mathrm{back}}$ denote the backbone features, $x_t^{(s_d)}$ the raw decoder patches, and $c$ the decoder condition. The resulting features are
\begin{equation}
    h=\mathcal D_\theta\!\left(
        \operatorname{split}(H_{\mathrm{in}}h^{\mathrm{back}})
        +X_{\mathrm{proj}}x_t^{(s_d)},\ c\right),
    \label{eq:decoder}
\end{equation}
where $\mathcal D_\theta$ denotes the decoder Transformer and $c$ is the projected sum of time and class embeddings. In our implementation, $s_b=16$, $s_d=8$, and $\mathcal D_\theta$ contains two blocks. The finer sequence is processed only by the decoder, while the deep backbone retains its original patch resolution. Each raw RGB decoder patch contributes $3s_d^2$ coordinates, which must be processed alongside the backbone features within the decoder's hidden representation.

\begin{table}[!ht]
    \centering
    \small
    \setlength{\tabcolsep}{5pt}
    \begin{tabular}{lccccc}
    \toprule
    Intervention & Backbone width & Patch & Decoder width & Raw input & FID-50K ($\downarrow$) \\
    \midrule
    Vanilla B & 768 & 16 & -- & No & 149.07 \\
    Vanilla L & 1024 & 16 & -- & No & 33.43 \\
    Vanilla B, smaller patches & 768 & 8 & -- & No & 28.70 \\
    \midrule
    Two-block decoder & 768 & 16 & 128 & Yes & 220.78 \\
    Two-block decoder & 768 & 16 & 256 & Yes & 34.68 \\
    Two-block decoder & 768 & 16 & 384 & No & 186.16 \\
    Two-block decoder & 768 & 16 & 384 & Yes & 31.85 \\
    \bottomrule
    \end{tabular}
    \vspace{4mm}
    \caption{\textbf{Interventions on bandwidth and raw access.} All rows use $v$-prediction, 250K updates, and no guidance. Vanilla models have no low-rank input bottleneck. Decoder models use a rank-128 backbone input bottleneck and patch-8 decoder tokens, without an output-side shortcut. ``Raw input'' denotes an additional decoder input, not the ordinary backbone input.}
    \label{tab:bandwidth}
    \vspace{-6mm}
\end{table}

As shown in Table~\ref{tab:bandwidth}, increasing backbone capacity or reducing patch size can restore vanilla $v$-prediction, but both interventions introduce additional computational costs.  In particular, a patch-8 backbone increases the sequence length throughout the network, whereas RiT processes finer-resolution tokens using only two lightweight decoder blocks. Within the decoder family, two width-384 blocks fail without direct raw-input access but converge when it is provided. Conversely, a width-128 decoder fails despite receiving the 192-dimensional raw subpatches, while widths 256 and 384 train effectively. These controlled interventions suggest that both raw-input availability and sufficient post-injection bandwidth are important for restoring $v$-prediction.

\vspace{-2mm}
\paragraph{A shared-output transport shortcut.}
The nonlinear decoder must still learn the simple raw-state contribution identified by the affine analysis. We provide a short output-side path for this term and combine it with semantic features before a shared output projection. For decoder width $d$, a SiLU followed by a linear layer maps the condition $c$ to five channel-wise coefficients:
\begin{equation}
    [c_1,c_2,c_3,c_4,c_5]
    =\operatorname{Linear}(\operatorname{SiLU}(c)),\qquad c_i\in\mathbb R^d.
    \label{eq:coefficients}
\end{equation}
Following the coefficient numbering in the implementation, the raw and semantic features are
\begin{equation}
    r=X_{\mathrm{in}}x_t^{(s_d)},\qquad
    q=H_{\mathrm{out}}\!\left(\operatorname{RMSNorm}(h)\odot(1+c_4)+c_5\right).
    \label{eq:branches}
\end{equation}
The fractional form is inspired by the rational time dependence of the optimal affine coefficient, which satisfies $C_i(1)=1$ at the pure-noise endpoint. We anchor the raw gain to one there and adopt the same endpoint convention for the semantic gain:
\begin{equation}
    d_t=\max\{t+(1-t)c_2,\epsilon\},\qquad
    g_r=\frac{1+(1-t)c_1}{d_t},\qquad
    g_h=\frac{1+(1-t)c_3}{d_t},
    \label{eq:anchored}
\end{equation}
with $\epsilon=0.05$, giving the final prediction
\begin{equation}
    \hat v_\theta
    =\operatorname{unpatchify}\!\left(X_{\mathrm{out}}(g_r\odot r+g_h\odot q)\right).
    \label{eq:rit-final}
\end{equation}
All divisions and denominator clipping operations are applied element-wise, and coefficients are broadcast over subpatch positions. The projections are learned affine maps. $X_{\mathrm{in}}$ is distinct from the decoder injection $X_{\mathrm{proj}}$, whereas $X_{\mathrm{out}}$ is shared by the raw and semantic branches. $H_{\mathrm{out}}$ maps semantic features into the common output coordinates before their scaled contributions are combined.

The endpoint constraints apply to the modulation gains rather than the entire velocity field, since the projections and semantic contribution remain learnable. Neither PCA coordinates nor exact conditional variances are prescribed. We initialize $H_{\mathrm{out}}$ and the coefficient outputs to zero, initialize $X_{\mathrm{in}}$ with orthonormal columns, and initialize $X_{\mathrm{out}}$ with its transposed weight, using zero biases. For the final decoder widths, the initial predictor is therefore $x_t/\bar t$. The transpose relation is not enforced during training, and the nonlinear decoder learns alongside the shortcut without requiring a precomputed PCA basis.
We compare several output formulations in Table~\ref{tab:shortcut}. The basic shortcut uses directly learned gains, the free fractional variant omits the $(1-t)$ factors, and the separate-denominator variant predicts independent denominators for the raw and semantic branches.

\vspace{-1mm}
\begin{table}[!ht]
    \centering
    \setlength{\tabcolsep}{7pt}
    \begin{tabular}{lcc}
    \toprule
    Output shortcut & Objective & FID-50K ($\downarrow$) \\
    \midrule
    None (standard decoder readout) & $v$ & 31.85 \\
    Basic learned gains & $v$ & 28.27 \\
    Free fractional, shared denominator & $v$ & 28.33 \\
    Anchored fractional, separate denominators & $v$ & 28.51 \\
    Anchored fractional, shared denominator (RiT) & $v$ & \textbf{27.97} \\
    Anchored fractional, shared denominator & $x$ & 28.49 \\
    \bottomrule
    \end{tabular}
    \vspace{3mm}
   \caption{\textbf{Output-shortcut ablation.} All models use the B-scale backbone, a rank-128 input bottleneck, and a raw-conditioned two-block width-384 decoder. Results are measured after 250K updates without guidance. The output-side shortcut improves FID over the standard decoder readout, with the anchored fractional parameterization performing best among the tested variants. The $x$-prediction row uses the same architecture with the clean-data objective to compare the two prediction targets.}
    \label{tab:shortcut}
    \vspace{-6mm}
\end{table}

Their exact coefficient assignments are given in Appendix~\ref{appc:variants}. Even the basic shortcut improves FID over the standard decoder readout. The shared anchored form achieves the best FID among the tested variants, while additional denominator freedom does not improve this comparison. The same architecture also supports $x$-prediction, but direct $v$-prediction achieves lower FID under the common recipe. These results complement the bandwidth interventions by showing the benefit of a short learnable transport path alongside raw-conditioned nonlinear decoding.

\section{Additional Experiments}
\vspace{-1mm}
\paragraph{Input bottleneck and time sampling.}
We further investigate architectural and training hyperparameters using RiT-B, with results summarized in Table~\ref{tab:hyperparams}. For the input bottleneck, we vary the intermediate width of the linear backbone embedding from 128 to 768 while keeping the two-block, 384-dimensional decoder unchanged. Reducing the bottleneck width consistently improves FID, with the 128-dimensional setting achieving the best result. Unlike the PCA intervention in Section~\ref{sec:pca}, these experiments retain the full data and noise distributions and constrain only the learned backbone input map. The bottleneck therefore affects only the backbone embedding, leaving the decoder's separate linear projection of the full raw input unchanged.

Regarding time sampling, we draw $\operatorname{logit}(t)\sim\mathcal N(\mu_t,\sigma_t^2)$ and evaluate four choices of $(\mu_t,\sigma_t)$. Increasing $\mu_t$ shifts the sampling distribution toward higher noise levels and improves FID. Further reducing $\sigma_t$ from 1.0 to 0.8 at $\mu_t=0.8$ yields the best result. Together, these observations are consistent with the design choices in JiT~\citep{li2025back}: once the routing bottleneck is alleviated, direct $v$-prediction also benefits from a low-rank input embedding and a noise-biased training distribution.

\begin{table}[ht]
\vspace{0mm}
  \centering
  \begin{minipage}{0.48\textwidth}
    \centering
    \setlength{\tabcolsep}{4pt}
    \small
    \begin{tabular}{c cccc}
    \toprule
    Bottleneck Width & 128 & 256 & 384 & 768 \\
    \midrule
    FID-50K ($\downarrow$) & \textbf{27.97} & 29.63 & 29.70 & 30.45 \\
    \bottomrule
    \end{tabular}
  \end{minipage}
  \hfill
  \begin{minipage}{0.50\textwidth}
    \centering
    \setlength{\tabcolsep}{2pt}
    \small
    \begin{tabular}{c cccc}
    \toprule
    $(\mu_t, \sigma_t)$ & (0.0, 1.0) & (0.4, 1.0) & (0.8, 1.0) & (0.8, 0.8) \\
    \midrule
    FID-50K ($\downarrow$) & 33.69 & 29.71 & 29.48 & \textbf{27.97} \\
    \bottomrule
    \end{tabular}
  \end{minipage}
  \vspace{4mm}
  \caption{\label{tab:hyperparams}\textbf{Hyper-parameter ablation.} Following the $x$-prediction baseline~\citep{li2025back}, we find that a lower-rank input bottleneck and a high-noise sampling bias consistently enhance performance. All results are obtained with RiT-B after 200 epochs without classifier-free guidance.}
  \vspace{-8mm}
\end{table}

\paragraph{Generation quality across model scales.}
To evaluate scalability, we train RiT-B/16 and RiT-L/16 for approximately 600 epochs and RiT-H/16 for approximately 480 epochs. These models build on the corresponding JiT Transformer backbones with our lightweight raw-conditioned decoders. Detailed training and inference configurations are provided in Appendix~\ref{appc:details}, including the late-stage shortcut learning-rate reduction for RiT-H. All models operate directly in pixel space without PCA projection or auxiliary perceptual, adversarial, frequency-domain, or representation-alignment losses.

Table~\ref{tab:maintable} summarizes the generative performance on ImageNet $256\times256$. RiT-B/16 and RiT-L/16 achieve FID scores of 3.01 and 2.06 after 600 epochs, compared with the reported JiT results of 3.66 and 2.36. RiT-H/16 reaches 1.79 after 480 epochs, while JiT-H/16 reports 1.86 after 600 epochs. These results support our analysis that $v$-prediction need not be inferior to $x$-prediction in pixel space. With sufficient bandwidth and direct raw-input access, $v$-prediction becomes effective across model scales. The previously observed failure therefore reflects a mismatch between the objective's information requirements and the architecture, not an inherent limitation of velocity prediction.

\newcolumntype{Y}{>{\centering\arraybackslash}X}
\begin{table*}[ht]
\centering
\setlength{\tabcolsep}{4pt}
\begin{adjustbox}{max width=1.\linewidth}
\begin{tabularx}{\linewidth}{l Y Y Y Y}
\toprule
Method & Epochs & Params & IS ($\uparrow$) & FID-50K ($\downarrow$) \\
\hline 
\rowcolor{gray!10!blue!5}\multicolumn{5}{l}{\textbf{\textit{Generative Adversarial Networks}}}\\
BigGAN~\citep{brock2019large} & - & 112M & 171.4 & 6.95 \\
StyleGAN-XL~\citep{sauer2022styleganxl} & - & 166M  & \textbf{265.1} & \textbf{2.30} \\
GigaGAN~\citep{kang2023scaling} & - & 569M & 225.5 & 3.45 \\
\hline
\rowcolor{gray!10!blue!5}\multicolumn{5}{l}{\textbf{\textit{Masked and Autoregressive Models}}}\\
LlamaGen-XL~\citep{sun2024llamagen}  & 300 & 775M & 244.1 & 2.62 \\
VAR-\textit{d30}~\citep{tian2024var} & 350  & 2.0B & \textbf{323.1} & 1.92 \\
MAR~\citep{li2024mar} & 800 & 943M & 303.7 & \textbf{1.55} \\ 
RandAR-XL~\citep{pang2025randar} & 300 & 775M & 314.2 & 2.22 \\
\hline
\rowcolor{gray!10!blue!5}\multicolumn{5}{l}{\textbf{\textit{Latent Diffusion Models}}} \\
LDM-4-G~\citep{rombach2022high} & 170 & 400M & 247.7 & 3.60 \\
DiT-XL/2~\citep{peebles2023scalable} & 1400 & 675M & 278.2 & 2.27 \\
SiT-XL/2~\citep{ma2024sit} & 1400 & 675M & 277.5 & 2.06 \\
LightningDiT-XL~\citep{yao2025reconstruction} & 800 & 675M & 295.3 & 1.35\\
REPA~\citep{yu2025representation}, SiT-XL/2 & 800 & 675M & \textbf{305.7} & 1.42 \\
RAE~\citep{zheng2025diffusion}, DiT$^{\mathrm{DH}}$-XL & 800 & 839M & 262.6 & \textbf{1.13} \\
\hline
\rowcolor{gray!10!blue!5}\multicolumn{5}{l}{\textbf{\textit{Pixel Space Diffusion Models}}} \\
ADM-G~\citep{dhariwal2021diffusion} & 400 & 554M & 186.7 & 4.59\\
RIN~\citep{jabri2023scalable} & 480 & 410M & 182.0 & 3.42\\
PixelFlow-XL/4~\citep{chen2025pixelflow} & 320 & 677M & 282.1 & 1.98 \\
PixNerd-XL/16~\citep{wang2025pixnerd}  & 160 & 700M & 297.0 & 2.15 \\
DiP-XL/16~\citep{chen2025dip} & 600 & 631M & 281.9 & 1.79\\
DeCo-XL/16$^\dagger$~\citep{ma2025deco} & 600 & 682M & \textbf{304.0} & \textbf{1.69} \\
\midrule
JiT-B/16~\citep{li2025back} & 600 & 131M & 275.1 & 3.66 \\
RiT-B/16 & 600 & 137M & 300.3 & 3.01 \\
JiT-L/16~\citep{li2025back} & 600 & 459M & 298.5 & 2.36 \\
RiT-L/16 & 600 & 470M & 331.4 & 2.06 \\
JiT-H/16~\citep{li2025back} & 600 & 953M & 303.4 & 1.86 \\
RiT-H/16 & 480 & 979M & \textbf{332.9} & \textbf{1.79}\\
\bottomrule
\end{tabularx}
\end{adjustbox}
\vspace{1mm}
\caption{\label{tab:maintable}\textbf{Class-conditional ImageNet $256\times256$ results.} Baselines follow their own reported sampling settings. $^\dagger$DeCo additionally employs frequency-aware and representation-alignment losses.}
\vspace{-4mm}
\end{table*}

\vspace{-1mm}
\paragraph{Performance at 200 epochs.}
The improvement is evident early in training. At 250K updates, corresponding to approximately 200 epochs, RiT-B/16, RiT-L/16, and RiT-H/16 achieve FID scores of 3.40, 2.48, and 2.17, respectively. These results improve upon the 4.37, 2.79, and 2.29 reported by JiT for the corresponding backbone scales at 200 epochs~\citep{li2025back}. Detailed EMA and guidance configurations for these intermediate checkpoints and the final models are provided in Appendix~\ref{appc:inference}.

Our contribution is primarily explanatory and methodological. We focus on understanding the information requirements of pixel-space prediction objectives and providing an architectural intervention that alleviates the resulting bandwidth bottleneck. This perspective complements other advances in generative modeling, including representation alignment (REPA~\citep{yu2025representation}), in-patch convolutional decoders (DiP~\citep{chen2025dip}), and frequency-aware training losses (DeCo~\citep{ma2025deco}). RiT provides a lightweight realization of the design principles identified by our analysis, rather than prescribing a single architecture for pixel-space generation. These results suggest ways to build on these principles through more expressive raw-conditioned decoders and complementary training techniques.

\vspace{-1mm}

\section{Conclusion}
We investigate the failure of direct $v$-prediction in pixel-space Flow Matching through patch-wise PCA, affine analysis, linear reconstruction probes, and controlled architectural experiments. Our results identify a bandwidth bottleneck in which preserving raw noisy state for minor-component velocity estimation competes with principal-component modeling. We further examine two complementary aspects of $x$-prediction: its parameterization provides a raw-input bypass, while its learned input embedding can spontaneously concentrate on principal directions. Under low-rank observations, our flow analysis characterizes $x$-prediction as an implicit conditional-mean decoder, completing discarded components without propagating their noisy states through the backbone. This completion can capture nonlinear dependence on retained information but does not generally recover the full conditional distribution. RiT combines a lightweight raw-conditioned nonlinear decoder for modeling conditional variation with a learnable output shortcut for raw-state transport, achieving competitive performance on ImageNet $256\times256$. These findings show that $v$-prediction can be effective when architectural bandwidth and information paths accommodate the demands of the prediction objective.

\section*{Acknowledgment and Funding Transparency Statement}
This work was supported by the Shanghai Qi Zhi Institute Innovation Program (SQZ202410). We gratefully acknowledge the Shanghai Qi Zhi Institute for providing access to NVIDIA A800 GPUs for our experiments. Beyond the institutional affiliations and financial and computational support disclosed in this paper, the authors have no additional competing interests to declare.

\bibliographystyle{plain}  
\bibliography{references}  


\appendix
\newpage
{\LARGE\bfseries Appendix}
\section{Mathematical Analysis of the Velocity Field in Minor Components}
\label{appa:proof}
In this section, we derive the optimal time-dependent affine coefficient $C_i(t)$ and its residual standard deviation, proving Proposition~\ref{prop:affine}. The calculation retains a nonzero patch mean and formalizes the raw-state transport required in noise-dominated directions. It explains the usefulness of the $x_t/t$ term in $x$-prediction and motivates learning this contribution through an explicit shortcut.

\subsection{Statistical Assumptions and Problem Setup}
Let $x_0\in\mathbb R^p$ denote an image patch with mean $\mu$ and covariance $\Sigma$. Its orthonormal eigenvectors $v_1,\ldots,v_p$ have eigenvalues $\lambda_1\ge\cdots\ge\lambda_p\ge0$. Let $x_1\sim\mathcal N(0,I)$ be independent Gaussian noise and $x_t=(1-t)x_0+tx_1$. For a fixed $t>0$, define
\begin{equation}
    \mu_i=v_i^\top\mu,\qquad
    U_i=v_i^\top(x_0-\mu),\qquad N_i=v_i^\top x_1.
\end{equation}
Then $\mathbb E[U_i]=\mathbb E[N_i]=0$, and independence gives
\begin{equation}
    \mathbb E[U_i^2]=\lambda_i,\qquad
    \mathbb E[N_i^2]=1,\qquad
    \mathbb E[U_iN_i]=0.
\end{equation}
Using the centered state $\widetilde x_t=x_t-(1-t)\mu$, write
\begin{equation}
    Q_i(t)=v_i^\top\widetilde x_t=(1-t)U_i+tN_i,\qquad T_i=N_i-U_i.
\end{equation}
The physical target coordinate is $v_i^\top(x_1-x_0)=T_i-\mu_i$. We seek the affine estimator $a_i v_i^\top x_t+b_i$ minimizing
\begin{equation}
    \mathcal L_i(a_i,b_i;t)
    =\mathbb E\left[\bigl(a_i v_i^\top x_t+b_i-v_i^\top(x_1-x_0)\bigr)^2\right].
\end{equation}
For any $a_i$, minimizing over the intercept gives $b_i^*=-\mu_i-(1-t)a_i\mu_i$. The remaining problem is therefore
\begin{equation}
    \mathcal L_i(a_i,b_i^*;t)=\mathbb E[(a_iQ_i(t)-T_i)^2].
\end{equation}
All expectations are taken at fixed $t$. Orthogonality separates the squared error along the axes; no independence between clean-data coordinates or Gaussian data assumption is required.

\subsection{Derivation of the Optimal Coefficient \texorpdfstring{$C_i(t)$}{C\_i(t)}}
Taking the derivative with respect to $a_i$ and setting it to zero yields
\begin{equation}
    2\mathbb E[(a_iQ_i-T_i)Q_i]=0,
    \qquad C_i(t)=\frac{\mathbb E[T_iQ_i]}{\mathbb E[Q_i^2]}.
\end{equation}
The denominator follows by expanding the noisy coordinate and using the vanishing cross term:
\begin{align}
    \mathbb E[Q_i^2]
    &=(1-t)^2\mathbb E[U_i^2]+2t(1-t)\mathbb E[U_iN_i]+t^2\mathbb E[N_i^2]\nonumber\\
    &=t^2+(1-t)^2\lambda_i.
\end{align}
Similarly, the numerator is
\begin{align}
    \mathbb E[T_iQ_i]
    &=\mathbb E[(N_i-U_i)((1-t)U_i+tN_i)]\nonumber\\
    &=t\mathbb E[N_i^2]-(1-t)\mathbb E[U_i^2]
      =t-(1-t)\lambda_i.
\end{align}
Since the denominator is positive for $t>0$, the optimal coefficient and the corresponding physical estimator are
\begin{equation}
    C_i(t)=\frac{t-(1-t)\lambda_i}{t^2+(1-t)^2\lambda_i},\qquad
    \widehat V_i^{\mathrm{aff}}(x_t,t)=-\mu_i+C_i(t)v_i^\top\widetilde x_t.
    \label{eq:app-affine}
\end{equation}
Thus, a nonzero mean changes the affine intercept without changing the covariance-dependent slope. This is the mean-corrected form of the component-wise transport analyzed in the main text.

\subsection{Derivation of the Residual Standard Deviation}
Define $r_i(t)=\widehat V_i^{\mathrm{aff}}(x_t,t)-v_i^\top(x_1-x_0)=C_i(t)Q_i-T_i$. Its mean is zero. Expanding its variance and substituting the optimal coefficient gives
\begin{align}
    \operatorname{Var}(r_i)
    &=C_i^2\mathbb E[Q_i^2]-2C_i\mathbb E[T_iQ_i]+\mathbb E[T_i^2]\nonumber\\
    &=\mathbb E[T_i^2]-\frac{\mathbb E[T_iQ_i]^2}{\mathbb E[Q_i^2]}.
\end{align}
Independence implies $\mathbb E[T_i^2]=1+\lambda_i$. Substituting the moments derived above and collecting terms yields
\begin{align}
    \operatorname{Var}(r_i)
    &=(1+\lambda_i)-\frac{(t-(1-t)\lambda_i)^2}{t^2+(1-t)^2\lambda_i}\nonumber\\
    &=\frac{(1+\lambda_i)(t^2+(1-t)^2\lambda_i)-(t-(1-t)\lambda_i)^2}
            {t^2+(1-t)^2\lambda_i}\nonumber\\
    &=\frac{\lambda_i\bigl(t^2+2t(1-t)+(1-t)^2\bigr)}
            {t^2+(1-t)^2\lambda_i}
     =\frac{\lambda_i}{t^2+(1-t)^2\lambda_i}.
\end{align}
Taking the square root gives the residual standard deviation in Proposition~\ref{prop:affine}:
\begin{equation}
    \operatorname{std}(r_i(t))
    =\frac{\sqrt{\lambda_i}}{\sqrt{t^2+(1-t)^2\lambda_i}}.
    \label{eq:app-residual}
\end{equation}
The variance explained by the affine estimate is
\begin{equation}
    \operatorname{Var}(\widehat V_i^{\mathrm{aff}})
    =\frac{(t-(1-t)\lambda_i)^2}{t^2+(1-t)^2\lambda_i}.
\end{equation}
The estimate and its residual are uncorrelated by the least-squares normal equation, and their variances sum to $1+\lambda_i$. The residual is measured against the paired training target; it may include irreducible conditional uncertainty and should not be interpreted entirely as a signal that nonlinear features can recover from the available noisy state.

\subsection{Implications for Model Architecture and Information Flow}
For fixed $t>0$, taking $\lambda_i\to0$ gives $C_i(t)\to1/t$ and $\operatorname{std}(r_i(t))\to0$. Restoring the mean, the estimated physical velocity approaches
\begin{equation}
    \widehat V_i^{\mathrm{aff}}(x_t,t)\longrightarrow
    -\mu_i+\frac{v_i^\top x_t-(1-t)\mu_i}{t}
    =\frac{v_i^\top x_t-\mu_i}{t}.
\end{equation}
Thus, in directions with sufficiently small data variance, a scaled raw-state term explains most of the target variation. The $x$-prediction parameterization supplies $x_t/t$ explicitly and uses its clean-data estimate to provide the remaining correction, including the nonzero mean. It can therefore reduce the need for the backbone to transmit each noisy coordinate through all its layers.

This argument concerns the linear contribution to the learning problem. The complete identity $(x_t-\mathbb E[x_0\mid x_t,t,\ell])/t$ remains valid for arbitrary data distributions at $t>0$; it does not assume that all eigenvalues vanish. The small-eigenvalue limit is also not uniform near the data endpoint: for every fixed $\lambda_i>0$, $C_i(t)\to-1$ as $t\to0$, while $C_i(1)=1$. This supports a transport prior without making $1/t$ the optimal affine slope throughout the flow.

Direct $v$-prediction must obtain the same state-dependent contribution from its available features. When noisy-state propagation competes with principal-component modeling, a raw-input shortcut can provide this contribution without routing it through the deep backbone. RiT learns the shortcut's coefficients jointly with a nonlinear decoder rather than prescribing a PCA basis or fixed covariance model. The additional ability of $x$-prediction to discard input coordinates and complete them nonlinearly is a separate mechanism, developed in Appendix~\ref{appb:lowrank}.

\section{Low-rank Observations and Conditional-mean Reconstruction}
\label{appb:lowrank}
In this section, we prove Proposition~\ref{prop:conditional-mean} using an integrating factor. The affine analysis in Appendix~\ref{appa:proof} explains the raw-state contribution to velocity estimation. Here, we examine how $x$-prediction handles information excluded by a low-rank input embedding. We retain the full data and noise distributions and allow nonlinear dependence between retained and discarded data components.

\subsection{Low-rank Observations with a Nonzero Mean}
Even with a full-width input embedding, Figure~\ref{fig:embedding} shows that $x$-prediction spontaneously learns low-effective-rank representations concentrated on leading PCA directions. This motivates our analysis of low-rank observations, although a low effective rank does not necessarily imply exact information loss. Moreover, JiT-B/16 and JiT-L/16 use an input embedding bottleneck of width 128, while JiT-H/16 uses width 256~\citep{li2025back}. Our final RiT models retain these bottlenecks. For two consecutive linear maps $W_1\in\mathbb R^{d_b\times p}$ and $W_2\in\mathbb R^{d_h\times d_b}$, their composition $W=W_2W_1$ satisfies
\begin{equation}
    \operatorname{rank}(W)\le d_b,\qquad
    \dim\ker W=p-\operatorname{rank}(W)\ge p-d_b.
    \label{eq:embedding-nullspace}
\end{equation}
For a 768-dimensional patch, at least 640 directions are therefore unobservable to the B/L backbones, and at least 512 to H. For any $\delta\in\ker W$, the embeddings of $x_t$ and $x_t+\delta$ are identical, so subsequent backbone layers cannot distinguish these inputs at fixed time and class. In a separate measurement of the rank-128 $x$-prediction model, over 99\% of the embedding operator energy lies in the leading 128 PCA directions after 250K training steps, providing further evidence of alignment with principal components. The discarded directions need not coincide exactly with individual PCA axes. Under the MSE objective, an optimal $x$-predictor must estimate their clean-data values from the retained information, leading to the conditional-mean analysis below.

For the PCA-aligned case in Proposition~\ref{prop:conditional-mean}, let $P_1=P_k$ and $P_2=I-P_1$. More generally, choose $P_1$ as the orthogonal projection onto the row space of the trained $W$, and $P_2$ onto its nullspace. The proof below applies to either choice. The observation restriction therefore applies without exact PCA alignment. The measured alignment explains why retained coordinates can be associated with principal directions in practice.
Define $\mu_j=P_j\mu$ and the centered data and noise components
\begin{equation}
    Y_0=P_1(x_0-\mu),\quad Z_0=P_2(x_0-\mu),\qquad
    Y_1=P_1x_1,\quad Z_1=P_2x_1.
\end{equation}
Then, with $\widetilde x_t=x_t-(1-t)\mu$, the noisy coordinates are
\begin{equation}
    Y_t=P_1\widetilde x_t=(1-t)Y_0+tY_1,\qquad
    Z_t=P_2\widetilde x_t=(1-t)Z_0+tZ_1.
    \label{eq:app-centered}
\end{equation}
For full images, the same notation denotes the corresponding block-diagonal projections across patches, the patch mean is repeated, and $Y_t$ includes all retained tokens. We assume finite data second moments and standard Gaussian noise $x_1\sim\mathcal N(0,I)$ independent of $(x_0,\ell)$, with time sampled independently.
For the projected noise, zero mean and $P_j^\top=P_j$, $P_j^2=P_j$, $P_1P_2=0$ give the following covariances:
\begin{align}
    \operatorname{Cov}(Y_1,Z_1)
    &=\mathbb E[(P_1x_1)(P_2x_1)^\top]
      =P_1\mathbb E[x_1x_1^\top]P_2^\top=P_1P_2=0,\nonumber\\
    \operatorname{Cov}(Z_1)
    &=\mathbb E[(P_2x_1)(P_2x_1)^\top]
      =P_2\mathbb E[x_1x_1^\top]P_2^\top=P_2.
    \label{eq:projected-noise-covariance}
\end{align}
Because $Y_1$ and $Z_1$ are jointly Gaussian, zero cross-covariance implies independence. Since the whole noise vector is independent of $(x_0,\ell)$, $Z_1$ is independent of $(Y_0,Z_0,Y_1,\ell)$ and hence of $(Y_t,t,\ell)$. In an orthonormal basis of the discarded subspace, $P_2$ represents identity covariance. The clean-data components $Y_0$ and $Z_0$ may nevertheless be dependent. Expanding the conditional covariance of $Z_t$ and using the same independence yields the exact relation
\begin{equation}
    \operatorname{Cov}(Z_t\mid Y_t,t,\ell)
    =(1-t)^2\operatorname{Cov}(Z_0\mid Y_t,t,\ell)+t^2P_2.
\end{equation}
Thus, at $t>0$, each discarded direction retains independent noise variance $t^2$ even if its clean value is predictable from the retained coordinates. This distinguishes loss of raw-state information from the possibility of predicting missing clean-data structure.
When $WP_2=0$, the embedding satisfies
\begin{equation}
    Wx_t+b=WY_t+(1-t)W\mu+b.
\end{equation}
If $W$ is injective on $\operatorname{range}(P_1)$, $WY_t$ uniquely determines $Y_t$. In particular, this is automatic when $P_1$ projects onto the row space: a vector in that space with $WY_t=0$ also lies in its orthogonal complement $\ker W$ and must be zero. The known time-dependent offset can be subtracted, so observing $(Wx_t+b,t,\ell)$ is equivalent to observing $(Y_t,t,\ell)$. This equivalence concerns the backbone input, and RiT's additional raw paths restore information outside it. 

Define the minor conditional mean
\begin{equation}
    m_t(y,\ell)=\mathbb E[Z_0\mid Y_t=y,t,\ell].
    \label{eq:conditional-mean}
\end{equation}
For any estimate $a$, representing a predicted value of the discarded clean-data component $P_2x_0$ based on the fixed observation $(y,t,\ell)$, subtracting and adding $\mu_2+m_t(y,\ell)$ in the error gives
\begin{align}
    &\mathbb E[\|a-P_2x_0\|_2^2\mid Y_t=y,t,\ell]\\
    &=\|a-\mu_2-m_t(y,\ell)\|_2^2-2(a-\mu_2-m_t(y,\ell))^\top
      \mathbb E[Z_0-m_t(y,\ell)\mid Y_t=y,t,\ell]\\
    &\quad+\mathbb E[\|Z_0-m_t(y,\ell)\|_2^2\mid Y_t=y,t,\ell]\\
    &=\|a-\mu_2-m_t(y,\ell)\|_2^2+\mathbb E[\|Z_0-m_t(y,\ell)\|_2^2\mid Y_t=y,t,\ell].
\end{align}
The cross term vanishes because $\mathbb E[Z_0-m_t(y,\ell)\mid Y_t=y,t,\ell]=0$ by the definition of the conditional mean. Only the first term depends on $a$, so the optimal minor clean-data estimate is $\mu_2+m_t(y,\ell)$. A positive time-dependent loss weight leaves this minimizer unchanged. Thus, a hard bottleneck guarantees an observation restriction, while the following endpoint result additionally assumes optimal regression and correctly learned retained dynamics.

\subsection{Integrating-factor Solution of the \texorpdfstring{$x$}{x}-Prediction Flow}
The optimal retained clean-data estimate is $\mu_1+\mathbb E[Y_0\mid Y_t,t,\ell]$. After subtracting the time-dependent mean, its unclipped velocity induces the principal equation
\begin{equation}
    y'(t)=\frac{y(t)-\mathbb E[Y_0\mid Y_t=y(t),t,\ell]}{t},\qquad y(1)=Y_1.
\end{equation}
As in Proposition~\ref{prop:conditional-mean}, assume this principal marginal flow is correctly learned and has a unique trajectory with an endpoint $y(0)$. Its right-hand side uses only the retained state, time, and class. The trajectory is therefore determined by $(Y_1,\ell)$ independently of $Z_1$, and need not follow an individual straight data-noise interpolation. Write the generated full trajectory as
\begin{equation}
    \widehat X(t)=(1-t)\mu+y(t)+z(t),\qquad
    y(t)\in\operatorname{range}(P_1),\quad z(t)\in\operatorname{range}(P_2),
\end{equation}
where $z(1)=Z_1$. Along this trajectory, abbreviate $m_t(y(t),\ell)$ as $m_t$. The minor part of the state is $u(t)=P_2\widehat X(t)=(1-t)\mu_2+z(t)$, and its predicted clean value is $\mu_2+m_t$. The velocity parameterization consequently gives
\begin{equation}
    u'(t)=\frac{u(t)-\mu_2-m_t}{t}
          =-\mu_2+\frac{z(t)-m_t}{t}.
\end{equation}
Differentiating the definition of $u(t)$ gives $u'(t)=-\mu_2+z'(t)$. Equating the two expressions cancels the mean terms and leaves
\begin{equation}
    z'(t)-\frac{z(t)}{t}=-\frac{m_t}{t}.
    \label{eq:app-minor-ode}
\end{equation}
Although $m_t$ depends nonlinearly on retained information, it is a prescribed function of time along $y(t)$. Equation~\eqref{eq:app-minor-ode} is consequently a linear ODE for $z(t)$.

Its integrating factor is $\exp(-\int t^{-1}\,\mathrm dt)=1/t$. Multiplication by this factor gives a product derivative, which can be integrated from $1$ to $t$:
\begin{align}
    \frac{\mathrm d}{\mathrm dt}\left(\frac{z(t)}{t}\right)
    &=\frac{z'(t)}{t}-\frac{z(t)}{t^2}=-\frac{m_t}{t^2},\nonumber\\
    \frac{z(t)}{t}-z(1)
    &=-\int_1^t\frac{m_s}{s^2}\,\mathrm ds
      =\int_t^1\frac{m_s}{s^2}\,\mathrm ds.
\end{align}
Multiplying by $t$ yields the explicit solution
\begin{equation}
    z(t)=tz(1)+t\int_t^1\frac{m_s}{s^2}\,\mathrm ds.
    \label{eq:minor-x-ode}
\end{equation}
We now evaluate the solution in Eq.~\eqref{eq:minor-x-ode} as $t\downarrow0$. The initial-noise term $tz(1)$ vanishes, so it remains to determine the limit of the integral term. Assume that $m_t$ is continuous along the principal trajectory and converges to $m_*=m_0(y(0),\ell)$ as $t\downarrow0$. Separating this limiting value from the integral gives
\begin{equation}
    t\int_t^1\frac{m_s}{s^2}\,\mathrm ds
    =(1-t)m_*+t\int_t^1\frac{m_s-m_*}{s^2}\,\mathrm ds.
\end{equation}
The first term converges to $m_*$. To show that the remaining term vanishes, fix any $\eta>0$. By continuity at $t=0$, there exists $\delta\in(0,1)$ such that $\|m_s-m_*\|<\eta/2$ whenever $0<s<\delta$. Fix this $\delta$ and define
\begin{equation}
    M_\delta=\int_\delta^1\frac{\|m_s-m_*\|}{s^2}\,\mathrm ds<\infty,
\end{equation}
which is finite by continuity on $[\delta,1]$. For any $0<t<\delta$, splitting the integral at $\delta$ gives
\begin{align}
    \left\|t\int_t^1\frac{m_s-m_*}{s^2}\,\mathrm ds\right\|
    &\le\frac{\eta}{2}t\int_t^\delta\frac{1}{s^2}\,\mathrm ds
       +t\int_\delta^1\frac{\|m_s-m_*\|}{s^2}\,\mathrm ds\nonumber\\
    &=\frac{\eta}{2}\left(1-\frac{t}{\delta}\right)+tM_\delta
    \le\frac{\eta}{2}+tM_\delta.
\end{align}
Now choose $0<t<\min\{\delta,\eta/[2(M_\delta+1)]\}$. The bound above is strictly smaller than $\eta$, proving that the remaining integral term converges to zero as $t\downarrow0$. Since $\eta$ was arbitrary, we obtain
\begin{equation}
    \lim_{t\downarrow0}t\int_t^1\frac{m_s}{s^2}\,\mathrm ds=m_*,
    \qquad z(0)=m_0(y(0),\ell).
\end{equation}
Restoring the nonzero mean gives
\begin{align}
    P_2\widehat X(0)
    &=\mu_2+m_0(y(0),\ell)\nonumber\\
    &=\mathbb E[P_2x_0\mid P_1x_0=\mu_1+y(0),\ell]\nonumber\\
    &=\mathbb E[P_2x_0\mid P_1x_0=P_1\widehat X(0),\ell].
    \label{eq:app-endpoint}
\end{align}
The second line follows from $Y_0=P_1x_0-\mu_1$, and the last uses $P_1\widehat X(0)=\mu_1+y(0)$. This proves Proposition~\ref{prop:conditional-mean}: under the stated assumptions, the discarded generated coordinates equal their conditional means given the retained endpoint and class, even when these means depend nonlinearly on the retained information.

\subsection{Comparison with Direct \texorpdfstring{$v$}{v}-Prediction and Implications for RiT}
We now examine what happens when a direct $v$-predictor is given the same low-rank observation $(Y_t,t,\ell)$. Since $Y_t=(1-t)Y_0+tY_1$, the velocity along the retained subspace satisfies $Y_1-Y_0=(Y_t-Y_0)/t$. Taking the conditional expectation gives
\begin{equation}
    \mathbb E[Y_1-Y_0\mid Y_t,t,\ell]
    =\frac{Y_t-\mathbb E[Y_0\mid Y_t,t,\ell]}{t}.
\end{equation}
This is exactly the retained velocity induced by the optimal $x$-predictor in the previous subsection. Thus, both objectives generate the same principal trajectory $y(t)$ under optimal regression, and we can compare their behavior in the discarded subspace along this trajectory.
The target velocity of the discarded components is $P_2(x_1-x_0)=Z_1-Z_0-\mu_2$. Its optimal MSE prediction is obtained by taking the conditional expectation:
\begin{align}
    \mathbb E[P_2(x_1-x_0)\mid Y_t,t,\ell]
    &=\mathbb E[Z_1\mid Y_t,t,\ell]
      -\mathbb E[Z_0\mid Y_t,t,\ell]-\mu_2\nonumber\\
    &=-m_t(Y_t,\ell)-\mu_2.
\end{align}
Here $\mathbb E[Z_1\mid Y_t,t,\ell]=0$ because the discarded Gaussian noise is independent of the retained observation, as established above. Along the principal trajectory, the generated minor state $u(t)=(1-t)\mu_2+z(t)$ therefore satisfies $u'(t)=-\mu_2-m_t$. Since $u'(t)=-\mu_2+z'(t)$, we obtain $z'(t)=-m_t$. Integrating backward from $t=1$ gives
\begin{equation}
    z(t)=z(1)+\int_t^1m_s\,\mathrm ds,
    \qquad z(0)=z(1)+\int_0^1m_s\,\mathrm ds.
    \label{eq:minor-v-endpoint}
\end{equation}
The integral depends only on the principal trajectory and class label, while $z(1)=Z_1$ is independent Gaussian noise. Consequently, conditioning on $y(\cdot)$ and $\ell$, the integral becomes fixed and the remaining randomness comes entirely from $Z_1$:
\begin{equation}
    \operatorname{Cov}(z(0)\mid y(\cdot),\ell)
    =\operatorname{Cov}(Z_1\mid y(\cdot),\ell)=P_2.
\end{equation}
The direct $v$-predictor can shift the initial noise according to the retained information, but its covariance remains unchanged. Without access to the discarded state, it cannot generally reproduce the conditional distribution of the clean-data minor components.

The two prediction objectives therefore produce different results under the same low-rank observation. As proved above, $x$-prediction removes the initial minor noise and reconstructs the centered discarded coordinates as $m_0(y(0),\ell)$. This allows the backbone to focus on retained information while completing the missing components through their conditional means. However, the reconstructed minor components are deterministic given the retained endpoint and class, whereas the original data may contain additional conditional variation. Direct $v$-prediction instead preserves the discarded Gaussian noise with unit variance and cannot adapt its distribution without observing that noise.

These results motivate the two components of RiT. By directly receiving the full noisy state, the raw-conditioned nonlinear decoder can process information discarded by the backbone bottleneck, allowing the backbone to focus on retained information while the decoder models nonlinear dependencies and conditional variation in minor components. This provides greater expressive capacity than the conditional-mean completion of the restricted $x$-predictor. The output-side shortcut further provides a learnable approximation to the affine transport analyzed in Appendix~\ref{appa:proof}, reducing the need for the decoder to learn this simpler contribution through its nonlinear layers. Together, these components allow RiT to retain a compact backbone representation while modeling richer joint conditional distributions in pixel space.

\subsection{Effect of the Time-denominator Cutoff}
The preceding analysis shows that the discarded components converge to their conditional means under the unclipped $x$-prediction flow. In practice, the velocity denominator is clipped as $\bar t=\max(t,\epsilon)$ to improve numerical stability near $t=0$. We now examine how this modification affects the endpoint behavior. The clipped and unclipped equations coincide for $t\ge\epsilon$. To isolate the final interval, we hold the principal trajectory fixed and let $M_t=\mu_2+m_t$. The clipped minor state $u_\epsilon(t)$ starts from $u_\epsilon(\epsilon)=u(\epsilon)$ and satisfies
\begin{equation}
    u_\epsilon'(t)-\frac{u_\epsilon(t)}{\epsilon}=-\frac{M_t}{\epsilon},
    \qquad 0\le t\le\epsilon.
\end{equation}
Multiplication by $e^{-t/\epsilon}$ gives
\begin{equation}
    \frac{\mathrm d}{\mathrm dt}\left(e^{-t/\epsilon}u_\epsilon(t)\right)
    =-\frac1\epsilon e^{-t/\epsilon}M_t.
\end{equation}
Integrating over $[0,\epsilon]$ yields
\begin{equation}
    e^{-1}u(\epsilon)-u_\epsilon(0)
    =-\frac1\epsilon\int_0^\epsilon e^{-s/\epsilon}M_s\,\mathrm ds,
\end{equation}
and consequently
\begin{equation}
    u_\epsilon(0)=e^{-1}u(\epsilon)
       +\frac1\epsilon\int_0^\epsilon e^{-s/\epsilon}M_s\,\mathrm ds.
    \label{eq:clipped-endpoint}
\end{equation}
Because $\epsilon^{-1}\int_0^\epsilon e^{-s/\epsilon}\,\mathrm ds=1-e^{-1}$, subtracting $M_0=\mu_2+m_*$ gives
\begin{align}
    u_\epsilon(0)-M_0
    &=e^{-1}\bigl(u(\epsilon)-M_0\bigr)
      +\frac1\epsilon\int_0^\epsilon e^{-s/\epsilon}(m_s-m_*)\,\mathrm ds,\nonumber\\
    \|u_\epsilon(0)-M_0\|
    &\le e^{-1}\|u(\epsilon)-M_0\|
       +(1-e^{-1})\sup_{0\le s\le\epsilon}\|m_s-m_*\|.
    \label{eq:clipping-bound}
\end{align}
The first term vanishes as $\epsilon\downarrow0$ by the endpoint proof, and the second vanishes by continuity. Small clipping gives a nearby solution when the conditional mean varies regularly near the endpoint, with its effect controlled by the two terms in this bound.
To track the remaining noise, substitute $t=\epsilon$ into Eq.~\eqref{eq:minor-x-ode} and restore the mean:
\begin{equation}
    u(\epsilon)=(1-\epsilon)\mu_2+\epsilon Z_1
       +\epsilon\int_\epsilon^1\frac{m_s}{s^2}\,\mathrm ds.
\end{equation}
Only the term $\epsilon Z_1$ depends on the discarded initial noise. Equation~\eqref{eq:clipped-endpoint} multiplies it by $e^{-1}$, leaving $\epsilon e^{-1}Z_1$ at the endpoint. At $\epsilon=0.05$, the residual noise standard deviation is approximately $0.0184$ in each orthonormal minor coordinate, compared with one under restricted direct $v$-prediction. Thus, when the conditional mean varies smoothly near $t=0$, the clipped solution remains concentrated near the unclipped conditional-mean endpoint, with only a small amount of residual noise. This analysis isolates the effect of the cutoff along a fixed principal trajectory, while numerical integration, regression error, and changes to the principal trajectory introduce additional approximation effects.

\subsection{A Visual Illustration of Principal-component Reconstruction}
\label{appb:pca}
\begin{figure}[ht]
    \centering
    \includegraphics[width=0.62\linewidth]{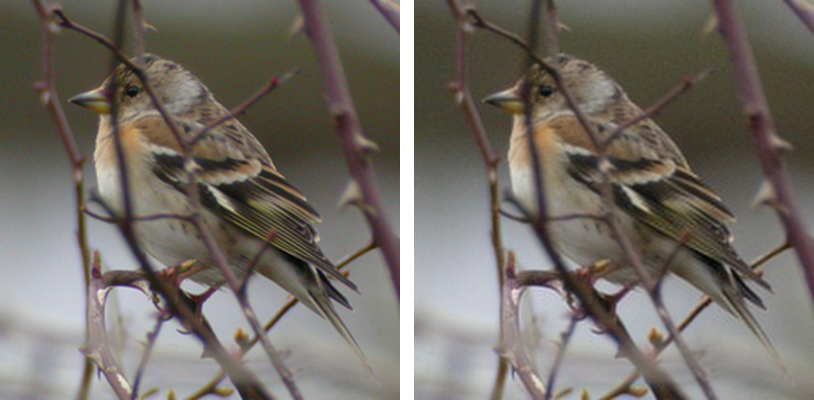}
    \caption{\textbf{Principal-component reconstruction.} Original image (left) and its reconstruction using the leading 128 of 768 principal components (right). Much of the visible structure is preserved, illustrating how retained principal components can provide sufficient information for conditional completion, although some fine details may remain ambiguous.}
    \label{fig:pca_img}
\end{figure}

We compare an image with its affine reconstruction using the leading 128 of 768 patch components in Figure~\ref{fig:pca_img}. Although many coordinates are discarded, much of the visible structure remains. This illustrates how the retained components can provide context for completing missing details, making conditional-mean reconstruction a plausible use of a compact representation. The displayed reconstruction is a linear PCA projection rather than the conditional-mean decoder analyzed above; it does not establish that discarded details are uniquely determined by the retained components.

\section{Implementation Details}
\label{appc:details}

\subsection{Architecture and Optimization}
The backbones follow the B, L, and H configurations of JiT~\citep{li2025back}. The principal architectural settings are listed in Table~\ref{tab:architecture}. The input embedding consists of two linear maps with the indicated intermediate width, which constrains the learned backbone input map without changing the pixel-space data or noise. We use fixed two-dimensional sinusoidal input position embeddings, two-dimensional rotary position embeddings in attention, RMS normalization, and SwiGLU feed-forward layers, with the same block type in the lightweight decoder. Thirty-two class tokens are inserted after 4, 8, and 10 backbone blocks for B, L, and H, respectively. These tokens participate in the remaining backbone blocks and are removed before the image features enter the decoder.

\begin{table}[ht]
    \centering
    \setlength{\tabcolsep}{9pt}
    \begin{tabular}{lccc}
    \toprule
    Setting & RiT-B/16 & RiT-L/16 & RiT-H/16 \\
    \midrule
    Image resolution & $256\times256$ & $256\times256$ & $256\times256$ \\
    Backbone patch size & 16 & 16 & 16 \\
    Backbone depth & 12 & 24 & 32 \\
    Backbone hidden width & 768 & 1024 & 1280 \\
    Backbone attention heads & 12 & 16 & 16 \\
    Input bottleneck width & 128 & 128 & 256 \\
    In-context class tokens & 32 & 32 & 32 \\
    Blocks before class-token insertion & 4 & 8 & 10 \\
    Decoder patch size & 8 & 8 & 8 \\
    Decoder depth & 2 & 2 & 2 \\
    Decoder hidden width & 384 & 512 & 768 \\
    Decoder attention heads & 6 & 8 & 12 \\
    Training updates & 750K & 750K & 600K \\
    Training epochs & 600 & 600 & 480 \\
    \bottomrule
    \end{tabular}
    \vspace{3mm}
    \caption{\textbf{Final model configurations.} All final models use full pixel-space data and noise, a raw-conditioned decoder, and the shared-denominator anchored shortcut. Epoch counts are approximate. PCA projection is used only in the diagnostic experiments, rather than in the final models.}
    \label{tab:architecture}
    \vspace{-5mm}
\end{table}

Training images are resized and center-cropped to $256\times256$, randomly flipped horizontally with probability 0.5, and normalized to $[-1,1]$. We use AdamW with global batch size 1024, betas $(0.9,0.95)$, and zero weight decay. The learning rate increases linearly to $2\times10^{-4}$ over the first 6250 updates and remains constant thereafter for B and L. We sample $\operatorname{logit}(t)\sim\mathcal N(0.8,0.8^2)$ unless explicitly varying the time distribution. The class condition is dropped with probability 0.1 to train the unconditional branch used by CFG, while attention and projection dropout are both zero. RiT-B is trained on 8 NVIDIA A800 GPUs with 80\,GB of memory per GPU. RiT-L and RiT-H each use 16 NVIDIA A800 GPUs with the same memory capacity, while retaining the global batch size of 1024.

For RiT-H, we multiply the effective learning rate of the raw shortcut input/output projections and the parameters producing $c_1,c_2,c_3$ by $0.01$ after 375K updates, corresponding to approximately 300 epochs. The remaining parameters, including the semantic projection and the parameters producing $c_4,c_5$, retain the base learning rate. This is a scale-specific stabilization choice rather than a change to the architecture or training objective, and it is not used for the 250K ablations or the B/L runs. We maintain EMA models with decay rates 0.9996, 0.9998, and 0.9999 and compare their sampling performance. The reported configurations in Table~\ref{tab:inference} use 0.9996, selected empirically from the evaluated EMA settings.

The backbone and decoder residual modulation layers are zero-initialized following the baseline convention. For shortcut variants, the final semantic projection and coefficient-prediction outputs are also zero-initialized. The raw shortcut input projection is initialized with orthonormal columns, and its output projection is initialized with the transposed weight. Both projections have zero-initialized biases. The input projection used to inject raw patches into the decoder blocks is initialized separately and is not shared with the shortcut. These are initialization choices only, and the projection weights remain learnable without orthogonality constraints or transpose tying during training.

\subsection{Shortcut Variants and Controlled Comparisons}
\label{appc:variants}
Table~\ref{tab:shortcut-definitions} specifies the coefficient assignments of the output-shortcut variants. The decoder condition $c$ is the sum of time and class embeddings, projected to the decoder width. One SiLU--linear mapping produces four, five, five, or six channel vectors for the basic, free fractional, anchored fractional, or separate-denominator variant, respectively. The names $c_i$ refer to the ordered outputs within each variant, following the implementation. The symbols $\gamma$ and $\beta$ denote the corresponding coefficients $c_i$ in Table~\ref{tab:shortcut-definitions} and do not introduce additional modulation parameters.

For these shortcut variants, the branch features are
\begin{equation}
    r=X_{\mathrm{in}}x_t^{(s_d)},\qquad
    q=H_{\mathrm{out}}\!\left(\operatorname{RMSNorm}(h)\odot(1+\gamma)+\beta\right),
\end{equation}
and their common pixel-patch output is $X_{\mathrm{out}}(g_r\odot r+g_h\odot q)$. The semantic projection $H_{\mathrm{out}}$ acts before $g_h$, and the two branches are added before the shared output projection. For brevity, we define
\begin{equation}
    \begin{aligned}
        n_i&=1+(1-t)c_i, &
        d_i&=\max\{t+(1-t)c_i,\epsilon\},\\
        f_i&=\max\{t+c_i,\epsilon\}, &
        \epsilon&=0.05.
    \end{aligned}
\end{equation}
These quantities are channel-wise and shared across all decoder tokens, depending only on the time and class condition. The raw and semantic features $r$ and $q$ are computed for each token from the noisy input and decoder hidden states, respectively. The two raw-input projections $X_{\mathrm{proj}}$ and $X_{\mathrm{in}}$ remain distinct, as described in Section~\ref{sec:rit}.

\begin{table}[ht]
    \centering
    \setlength{\tabcolsep}{9pt}
    \begin{tabular}{lccc}
    \toprule
    Variant & $g_r$ & $g_h$ & $(\gamma,\beta)$ \\
    \midrule
    Basic & $1+c_1$ & $1+c_2$ & $(c_3,c_4)$ \\
    Free fractional & $(1+c_1)/f_2$ & $(1+c_3)/f_2$ & $(c_4,c_5)$ \\
    Anchored fractional & $n_1/d_2$ & $n_3/d_2$ & $(c_4,c_5)$ \\
    Anchored, separate denominators & $n_1/d_2$ & $n_3/d_4$ & $(c_5,c_6)$ \\
    \bottomrule
    \end{tabular}
    \vspace{3mm}
    \caption{\textbf{Coefficient assignments for the output-shortcut ablation.} All operations are channel-wise. The final RiT uses the anchored fractional variant with a shared denominator, while retaining separate learned numerators for the raw and semantic branches.}
    \label{tab:shortcut-definitions}
    \vspace{-3mm}
\end{table}

The no-shortcut variant instead produces two coefficients and directly predicts the pixel patch as $H_{\mathrm{out}}(\operatorname{RMSNorm}(h)\odot(1+c_1)+c_2)$. In this variant, $H_{\mathrm{out}}$ maps to the pixel-patch dimension rather than the decoder hidden width, and there is no additional $X_{\mathrm{in}}$ or $X_{\mathrm{out}}$ branch. It can still receive raw input inside the decoder through $X_{\mathrm{proj}}$. Thus, disabling decoder injection and removing the output shortcut are separate interventions. In Table~\ref{tab:bandwidth}, all decoder rows remove the output shortcut, and the width-384 ``No'' row additionally removes raw-input injection into the decoder blocks.

With the initialization described above, the no-shortcut readout initially outputs zero. For the tested shortcut models, the raw input/output composition is initially the identity, so the basic shortcut initially outputs $x_t$, whereas the fractional variants initially output $x_t/\max(t,\epsilon)$. These initial outputs are determined by the architecture and initialization, independently of the prediction objective. During training, the network output is supervised as either velocity or clean data, depending on whether $v$-prediction or $x$-prediction is used.

PCA experiments use $k\in\{128,192,256,384,768\}$, backbone width 768, embedding intermediate width 768, no added decoder blocks, and no output shortcut. Data and noise are projected by the same affine map in Eq.~\eqref{eq:affine-pca}. The vanilla-L intervention has hidden and embedding intermediate widths 1024, 24 blocks, and patch size 16. The vanilla patch-8 intervention keeps a 12-block width-768 backbone but reduces raw patch dimension to 192. Neither intervention imposes a low-rank input bottleneck. 
All these controls use the same global batch size of 1024 and are trained for 250K updates, allowing us to examine the effects of backbone capacity and raw patch dimensionality under the same training schedule. Their computational costs differ, since increasing backbone width or reducing patch size changes the amount of computation required per update.

For the decoder and shortcut ablations, the B backbone has a width-128 input bottleneck. Decoder widths 128, 256, and 384 use 2, 4, and 6 attention heads, respectively. The final B model is the shared-denominator anchored setting. Bottleneck and time-sampling studies vary only their named setting relative to this model. In the corresponding $x$-prediction shortcut experiment, the architecture and initialization are unchanged, but its output is trained as a clean-image prediction using Eq.~\eqref{eq:xloss} and converted to velocity during sampling. These comparisons use a common training recipe to isolate the effects of the tested architectural and training choices, without additional hyperparameter tuning for individual variants.

\subsection{Sampling and Evaluation}
\label{appc:inference}
We use 50 uniformly spaced integration steps from $t=1$ to $t=0$. The first 49 steps use the second-order Heun update, and the final step uses a single Euler update directly to the data endpoint. This avoids evaluating the velocity at $t=0$, where the unregularized $x$-prediction parameterization contains a singular $1/t$ factor. Our implementation also clips this denominator as described in the preliminary section; we retain the same final-step convention for both objectives. The resulting sampler requires 99 velocity-function evaluations over 50 integration steps. CFG evaluates conditional and unconditional predictions together and combines them as
\begin{equation}
    v_{\mathrm{CFG}}=v_{\varnothing}+w\,(v_\ell-v_{\varnothing}).
\end{equation}
Guidance is active only within the specified time interval, and we use $w=1$ outside that interval. Each guided evaluation computes both conditional and unconditional predictions in a doubled batch, requiring approximately twice the computation of an unguided evaluation. The reported 99 velocity-function evaluations count these combined predictions as one solver evaluation.
\begin{table}[ht]
    \centering
    \setlength{\tabcolsep}{9pt}
    \begin{tabular}{lccccc}
    \toprule
    Model & Updates & Epochs & EMA & CFG scale & CFG interval \\
    \midrule
    RiT-B/16 & 250K & 200 & 0.9996 & 3.1 & $[0.0,0.89]$ \\
    RiT-L/16 & 250K & 200 & 0.9996 & 2.5 & $[0.0,0.89]$ \\
    RiT-H/16 & 250K & 200 & 0.9996 & 2.4 & $[0.0,0.89]$ \\
    \midrule
    RiT-B/16 & 750K & 600 & 0.9996 & 2.8 & $[0.0,0.89]$ \\
    RiT-L/16 & 750K & 600 & 0.9996 & 2.6 & $[0.0,0.89]$ \\
    RiT-H/16 & 600K & 480 & 0.9996 & 2.35 & $[0.0,0.90]$ \\
    \bottomrule
    \end{tabular}
    \vspace{3mm}
    \caption{\textbf{Inference configurations for the reported guided results.} Epoch counts are approximated from training updates and batch size. All configurations use 50-step Heun sampling with a final Euler step. Unguided ablations use $w=1$ and EMA 0.9996 with the same integration procedure.}
    \label{tab:inference}
    \vspace{-4mm}
\end{table}

FID-50K and IS are computed on 50,000 class-conditional samples using the ADM evaluation implementation~\citep{dhariwal2021diffusion} and its ImageNet reference statistics. Generated images are mapped from $[-1,1]$ to 8-bit RGB by rounding and clipping. We select guidance settings separately for different model scales and training horizons. Table~\ref{tab:inference} summarizes the configurations used to obtain the reported results, while unguided ablations are evaluated with CFG disabled.

\subsection{Patch Statistics and Representation Probes}
The patch statistics are computed over the ImageNet training set after the same resize, crop, and intensity normalization as training. The empirical covariance is centered by the patch mean. We account for horizontal flips with probability 0.5, including the between-mean covariance term when combining original and flipped patch distributions. These statistics define the affine PCA intervention and diagnostic plots, but are not required by the final RiT models during either training or sampling.

The embedding analysis uses the ordinary training weights, not EMA, at the recorded checkpoints. We compose the two consecutive linear embedding weights as $W_2W_1$ and compute the effective rank from the singular-value entropy in double precision. The PCA energy ratio uses a fixed $K=280$ for every model and checkpoint, without eigenvalue weighting, so it measures the learned map's response to the same set of unit-norm PCA directions. The full-L curve corresponds to the vanilla-L control. In particular, the PCA-128 curves describe training with projected data and noise while retaining a full-width embedding. They should therefore be distinguished from the architectural input-bottleneck ablation, which keeps the full pixel-space distributions unchanged.

The reconstruction probes use the ordinary 250K weights in evaluation mode and a fixed set of 1024 training images, processed in 16 batches of 64. We construct noisy inputs using the same interpolation and time-sampling procedure as training, with Gaussian noise $x_1\sim\mathcal N(0,I)$ and $\operatorname{logit}(t)\sim\mathcal N(\mu_t,\sigma_t^2)$. The parameters $(\mu_t,\sigma_t)$ follow the corresponding training configuration, and the noisy state is formed as $x_t=(1-t)x_0+tx_1$. Image selection, horizontal flips, Gaussian noise draws, and the underlying standard-normal draws for the time samples are fixed across the compared models. For PCA runs, the same affine projection is applied to data and noise before forming the noisy input. The forward passes use full precision and retain the class labels without CFG dropout.

We collect image-token features after each backbone block, excluding the in-context class tokens. For each layer, we concatenate the features and corresponding noisy patches from all 16 batches, then fit a single affine least-squares regression in double precision. We compute the $R^2$ score on the same data used to fit the regression. This score measures how much of the noisy input $x_t$ can be linearly reconstructed from the hidden states at each layer.

\section{Limitations and Broader Impact}
\paragraph{Limitations.}
Our experiments focus on class-conditional ImageNet generation at $256\times256$ and B, L, and H model scales. Generalization to higher resolutions, text-conditioned generation, other data distributions, and larger models remains to be studied. RiT provides a simple and effective realization of the identified bandwidth principle, while more expressive and efficient raw-conditioned decoder designs may offer further improvements in future work.

\paragraph{Broader Impact.}
Our work provides insights into the relationship between prediction objectives and architectural information flow, which may facilitate the development of more effective end-to-end pixel-space generative models. As with other image generation methods, improved synthesis capabilities may also be misused to produce misleading content or deepfakes. Since this work focuses on fundamental modeling principles, we do not introduce additional deployment-specific safeguards, which remain important considerations for practical applications.

\end{document}